%% file: main.tex
\documentclass{article}

 \usepackage[preprint]{neurips_2026}

\usepackage[utf8]{inputenc} 
\usepackage[T1]{fontenc}    
\usepackage{hyperref}       
\usepackage{url}            
\usepackage{booktabs}       
\usepackage{amsfonts}       
\usepackage{nicefrac}       
\usepackage{microtype}      
\usepackage{xcolor}         

\usepackage{graphicx}
\usepackage{subcaption}

\usepackage{amsmath}
\usepackage{amssymb}
\usepackage{mathtools}
\usepackage{amsthm}

\usepackage[capitalize,noabbrev]{cleveref}

\usepackage{siunitx}
\usepackage{colortbl}%

\newcommand{\name}{{MuseVLA}} 

\newcommand{\piZero}{$\pi_0$}
\newcommand{\piZeroFive}{$\pi_{0.5}$}

\newcommand{\minorEdit}[1]{\textcolor{black}{#1}}
\newcommand{\majorEdit}[1]{\textcolor{black}{#1}}

\newcommand{\heading}[1]{\smallskip\noindent\textbf{#1}}

\title{\name{}: An Adaptive Multimodal Sensing Vision-Language-Action Model for Robotic Manipulation} 

\author{%
  \textbf{Xingyuming Liu}$^{1,2,*}$ \qquad
  \textbf{Ruichun Ma}$^{2,\dagger}$ \qquad
  \textbf{Heyu Guo}$^{3,*}$ \\[4pt]
  \textbf{Qixiu Li}$^{2,4,*}$ \qquad
  \textbf{Qingwen Yang}$^{1,2,*}$ \qquad
  \textbf{Lin Luo}$^{2}$ \\[4pt]
  \textbf{Shiqi Jiang}$^{2}$ \qquad
  \textbf{Chenren Xu}$^{1}$ \qquad
  \textbf{Jiaolong Yang}$^{2}$ \qquad
  \textbf{Baining Guo}$^{2}$ \\[8pt]
  $^{1}$School of Computer Science, Peking University \qquad
  $^{2}$Microsoft Research Asia \\[2pt]
  $^{3}$Princeton University \qquad
  $^{4}$Tsinghua University%
}

\begin{document}

\maketitle

\renewcommand{\thefootnote}{\fnsymbol{footnote}}
\footnotetext[1]{Work done during internship at Microsoft Research Asia.}
\footnotetext[2]{Corresponding author.}
\renewcommand{\thefootnote}{\arabic{footnote}}

\input{sections/abstract}

\input{sections/intro}

\input{sections/related}

\input{sections/data}

\input{sections/method}

\input{sections/eval}

\input{sections/conclusion}

\bibliography{biblio}
\bibliographystyle{plainnat}


\appendix



\input{sections/appendix}



\end{document}

%% file: sections/abstract.tex
\begin{abstract}

Humans naturally leverage diverse sensing modalities to interact with the physical world, 
while most Vision-Language-Action (VLA) models for robotics rely solely on RGB observations.
This limits their ability to perceive physical properties that are difficult or impossible to infer from RGB cameras, such as temperature, sound, or radar response.
We present \name{}, an adaptive multimodal sensing VLA model that integrates novel sensors as on-demand tools for robotic manipulation.
Given a task instruction and visual context, \name{} first generates a sensor token and target description that select the sensing modality to invoke and what to attend to, analogous to a tool call with arguments.
It then converts the selected sensor measurement into a \emph{grounded sensor image}, a unified intermediate representation that encodes heterogeneous readings for multimodal fusion and action generation.
This design decouples sensor-specific processing from the VLA backbone, enabling efficient integration of diverse modalities.
To reduce the need for expensive multisensory robot datasets, we further introduce a data synthesis pipeline that augments existing RGB video datasets with grounded sensor images, enabling generalization to unseen sensor-guided tasks.
We evaluate \name{} on a real-world robot across challenging dexterous hand manipulation tasks that require multimodal sensing inputs, including temperature-guided pick-and-place, audio-driven object search, and radar-assisted hidden object retrieval.
\name{} achieves 80.6\% success rate on average, outperforming RGB-only and multisensory VLA baselines significantly, and exhibits strong zero-shot capabilities on unseen tasks.
Code, model and dataset are available at \url{https://github.com/microsoft/MuseVLA}.


\end{abstract}



%% file: sections/intro.tex
\section{Introduction}

Humans perceive and interact with the physical world through a rich set of sensing modalities, e.g., vision, sound, touch.
Crucially, humans do not fuse all available senses at all times.
Instead, they treat sensory modalities as \emph{task-oriented tools}, flexibly incorporating diverse inputs and adaptively attending to those most relevant to the task at hand.
Moreover, humans continuously {scale} their sensing capabilities by developing and adopting external sensor tools and learning when and how to use them on demand.
This ability to adaptively select and invoke the right sensing tool in a task-conditioned manner is fundamental to how to efficiently act in complex physical environments.

Vision-Language-Action (VLA) models have emerged as a powerful paradigm for robotic manipulation by leveraging web-scale vision-language pretraining for strong generalization and instruction following~\citep{black2024pi0, kim2024openvla, mees2024octo,pmlr-pi05, shukor2025smolvla, li2025vitra,zhang2026clap}.
Most of these VLA models rely solely on RGB images as visual input, while robots can have access to heterogeneous non-visual sensory observations such as thermal, acoustic, or radar signals, capturing complementary physical information beyond vision.
Recent efforts have begun to incorporate additional modalities such as depth~\citep{li2025pointvla, bhat20253d, qu2025spatialvla, liu2025mla}, tactile~\citep{bi2025vla, huang2025tactile}, and audio~\citep{zhao2025vlas, jones2025FuSe}.
While these approaches show promising gains, they exhibit several key limitations.
First, they typically rely on specialized model architectures or encoders for each sensor modality, which is not applicable to a diverse and evolving set of sensors.
Second, they depend on large-scale sensory datasets for training, which are often expensive and time-consuming to collect.
Lastly, existing models typically assume static sensor usage, lacking task-conditioned, adaptive selection of sensory inputs for efficiency and performance.

\begin{figure*}[t]
    \centering
    \includegraphics[width=0.99\textwidth]{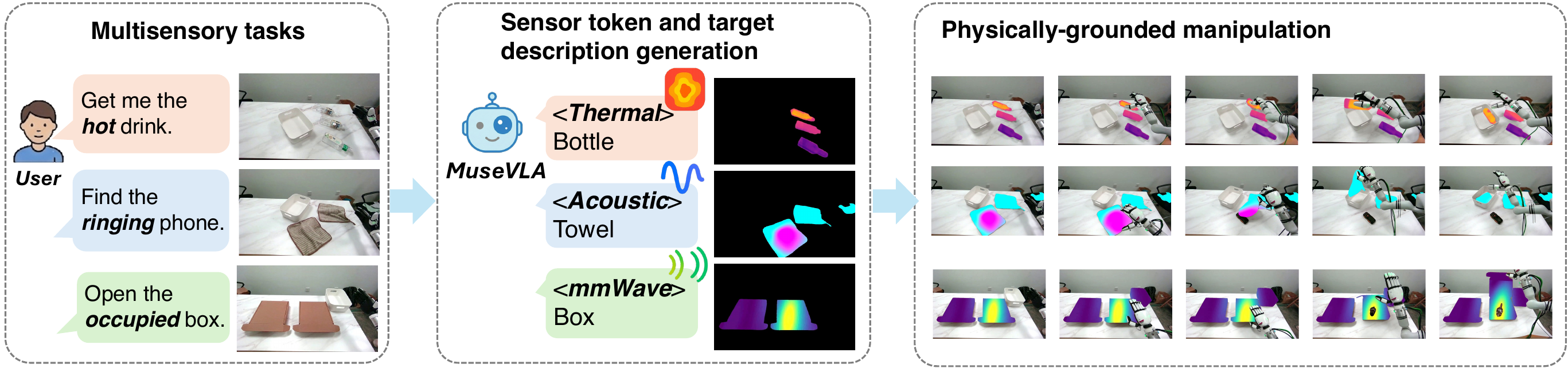}
    \caption{\textbf{Adaptive multisensory robotic manipulation.} \name{} targets manipulation tasks requiring multimodal sensing beyond RGB. It adaptively selects the suited sensor and generates a target description to construct a grounded sensor image that guides manipulation.}
    \label{fig:intro-overview}
\end{figure*}

In this work, we introduce \textbf{\name{}}, an adaptive multimodal-sensing VLA model that unifies diverse sensing modalities for robotic manipulation (\Cref{fig:intro-overview}).
\name{} presents a new paradigm for scaling VLA capability by treating sensors as on-demand tools, enabling conditional and scalable sensor perception within a single model.
\name{} dynamically selects a sensor tool based on task context, among thermal camera, mmWave radar, microphone, etc.
This allows the model to acquire task-oriented physical observations, which boosts performance and efficiency.
We demonstrate this on dexterous hand, a challenging setting where multisensory manipulation remains largely unexplored.

To achieve this, we introduce learnable \textit{sensor tokens} to enable adaptive sensor selection. 
Given task instruction and RGB observation, \name{} generates the selected sensor token together with a target description, analogous to invoking a tool with structured arguments.
Sensor observations are then processed and appended as input for subsequent action generation.
To support heterogeneous sensors without modality-specific encoders, we propose \textit{grounded sensor images} that spatially ground non-visual sensor responses onto the camera image plane over the target object region, enabling unified perception across diverse modalities using a pretrained vision encoder.
Building on this representation, we develop a data synthesis pipeline that generates multisensory training data from existing RGB datasets, enabling data-efficient training and strong zero-shot generalization.

We evaluate \name{} with a real-world multi-sensor robot setup, using a suite of challenging robotic manipulation tasks that require sensing modalities, including thermal-guided object pick and place, audio-driven object search, and mmWave radar-assisted hidden object retrieval, and their cascaded multi-stage tasks, each involving manipulating diverse objects with a dexterous hand.
Results show 80.6\% task success rate on average, significantly outperforming RGB-only baseline and raw sensor heatmaps based VLA baselines by 58\% and 47\% respectively.
\name{} also demonstrates effective adaptive sensor selection with high accuracy of generated sensor token and target description.
For unseen tasks, \name{} trained on synthesized dataset shows strong zero-shot generalization, achieving 66.7\% success rate on average.

To summarize, we make the following contributions:
\vspace{-8pt}
\begin{itemize}
    \item We present \name{}, an adaptive multimodal sensing VLA model that enables scalable and efficient sensor integration as on-demand tools for dexterous hand based manipulation. 
    \item We introduce sensor tokens for adaptive sensor selection and the unified grounded sensor image representation. We further propose a data synthesis pipeline that generates multisensory datasets from existing RGB robotic datasets for generalization to unseen tasks. 
    \item We evaluate \name{} on a suite of challenging robotic manipulation tasks and demonstrate its superior performance in leveraging sensing inputs.
    We will open-source our data and code to the community to facilitate further research.
\end{itemize}






%% file: sections/related.tex
\section{Related Works}


\heading{VLA models for robotic manipulation.}
Vision-language-action models process language instructions and visual observations to generate robot actions end-to-end~\citep{mees2024octo, kim2024openvla, wen2025dexvla, ze2024generalizable, black2024pi0, shukor2025smolvla}. 
Unlike conventional manipulation policies trained on specific tasks~\citep{chebotar2023q, luo2023action}, VLA models leverage web-scale pretraining to achieve strong generalization and instruction following.
Recent efforts have further scaled VLA models across diverse embodiments and large-scale data~\citep{intelligence2026pi, wu2026pragmatic, bjorck2025gr00t, generalist2026gen1}.
\minorEdit{Complementary to these scaling efforts, \name{} integrates sensing modalities as on-demand tools to empower VLA models for multisensory manipulation. In particular, we explore challenging dexterous hand manipulation tasks~\citep{li2025vitra,gao2026dreamdojo} that require fine-grained spatial understanding and precise control.}

\heading{Multi-sensor fusion.}
Fusing complementary sensors has proven critical for various domains, such as autonomous driving~\citep{liu2022bevfusion, wolters2024unleashing, Lin_2024_CVPR, zheng2025doracamom, palladin2024samfusion, li2025rctrans, kim2024crt, xiong2025lxlv2}, 3D scene understanding~\citep{Wang_2024_CVPR, han2025multimodal}. 
Sensing modalities provide richer physical information and more robust representations than any single camera alone.
Existing fusion methods are predominantly designed for perception tasks with sensor-specific architectures, rather than for general action generation and manipulation. 
\name{} presents a scalable and adaptive multi-sensor fusion approach within a unified VLA framework for manipulation.




\heading{Multisensory VLA models.}
Recent works extend VLA models with additional sensing modalities, including depth for spatial understanding~\citep{li2025pointvla, bhat20253d, qu2025spatialvla, patratskiy2025spatial, zhen20243d}, tactile feedback for force-aware control~\citep{bi2025vla, huang2025tactile, yu2025forcevla}, speech for human-robot interaction~\citep{zhao2025vlas}, and multisensory fusion~\citep{liu2025mla, guo2025omnivla}. 
These approaches show promising results, but they either target a specific modality with specialized architectures, or assume a fixed, predetermined sensor configuration with limited generalization to new tasks.
\name{} builds on and goes beyond these works by introducing learnable {sensor tokens} for task-conditioned adaptive sensor selection that boosts performance and efficiency, data synthesis for data-efficient training and generalization, and targeting \emph{dexterous hand} manipulation unexplored by prior work.


\begin{figure*}[t]
    \centering
    \includegraphics[width=0.95\textwidth]{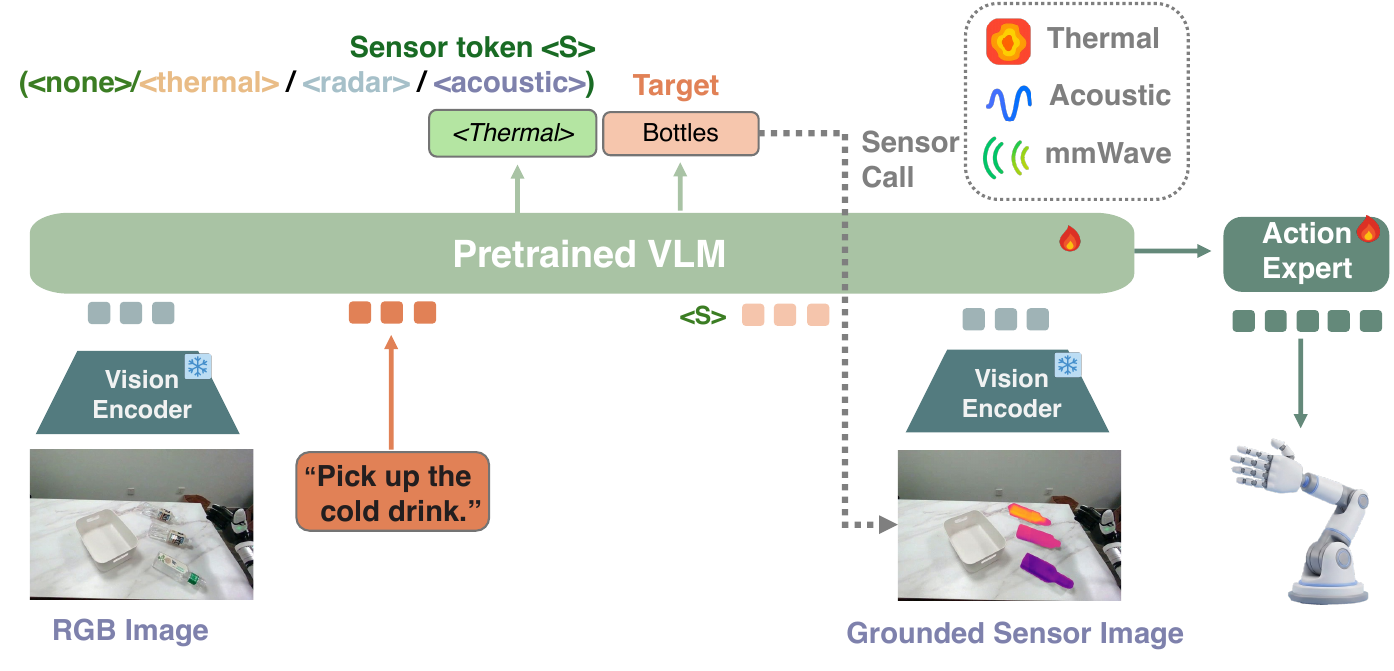}
    \caption{\textbf{Overview of \name{} model.} Given an RGB image and task instruction, \name{} generates a sensor token and target description. The selected sensor is invoked to construct a grounded sensor image, which is appended as input for manipulation action generation. We co-train VLM backbone and action expert end-to-end on real-world and synthesized multisensory datasets.}
    \label{fig:model-overview}
\end{figure*}

%% file: sections/data.tex
\section{Adaptive Multimodal Sensing}


\heading{Task definition.}
Given a manipulation task with language instruction $\boldsymbol{l}$,
the robot is equipped with sensors $\mathcal{S} = \{S_1, \ldots, S_N\}$, each providing observation $\boldsymbol{s}_i$.
The goal is to learn a VLA model $\boldsymbol{\pi}$ that maps RGB observations and sensory inputs to an action chunk $A = (\boldsymbol{a}_t, \ldots, \boldsymbol{a}_{t+H})$:
\begin{equation}
\boldsymbol{\pi}:\left(\boldsymbol{l},\; \boldsymbol{o}_t,\; \boldsymbol{s}_{1,t}, \dots, \boldsymbol{s}_{N,t}\right) \rightarrow A
\end{equation}
where $\boldsymbol{o}_t$ is the RGB observation and $\boldsymbol{s}_{i,t}$ is the observation from sensor $S_i$ at time $t$.
However, naively incorporating all sensor input is inefficient, as not all are relevant to every task. It also increases both training and inference computational cost and constrains the model to a fixed sensor set, limiting flexible model deployment across robotic platforms.

\heading{Decoupling sensors with VLA model.}
We introduce an adaptive sensor selection mechanism that decomposes the task into three stages: (1) the VLM model selects a task-relevant sensor $\boldsymbol{l_s}$ and a {sensing target description} $\boldsymbol{l_d}$; (2) the selected sensory observation is processed into an intermediate representation $\boldsymbol{m}$; and (3) this representation is autoregressively appended for action generation.
\begin{equation}
\boldsymbol{\pi}: (\boldsymbol{l}, \boldsymbol{o}_t) \rightarrow (\boldsymbol{l_s}, \boldsymbol{l_d}), \quad
\mathcal{G}: (\boldsymbol{o}_t, \boldsymbol{s}_{i,t}, \boldsymbol{l_d}) \rightarrow \boldsymbol{m}_{i,t}, \quad
\boldsymbol{\pi}: (\boldsymbol{l}, \boldsymbol{o}_t, \boldsymbol{l_s}, \boldsymbol{l_d}, \boldsymbol{m}_{i,t}) \rightarrow A
\label{equ:vla-process}
\end{equation}
Here, $\mathcal{G}$ is the sensor grounding function that constructs a grounded sensor image $\boldsymbol{m}_{i,t}$ from the RGB observation, the selected sensor's observation $\boldsymbol{s}_{i,t}$ (with $i$ specified by $\boldsymbol{l_s}$), and the target description.
The same VLA model $\boldsymbol{\pi}$ is reused for both sensor selection and action generation, maintaining a unified end-to-end architecture.
\minorEdit{By decoupling sensors from the VLA backbone via an intermediate representation, sensor modalities and backbone capability can be scaled independently, enabling flexible integration of new sensors without full retraining.}
It also brings efficient action generation since only the relevant sensor is processed.



\heading{Sensor data representation.}
\label{sec:masked-images}
We introduce \majorEdit{grounded sensor images} as the intermediate representation $\boldsymbol{m}$, which encodes diverse sensor data into RGB space by overlaying sensor heatmaps onto task-relevant regions (\Cref{fig:sensor-masked}).
This unifies sensing modalities without specialized encoders, inspired by how humans ground sensory physical information onto specific objects or regions in the RGB space. \majorEdit{This unified format also enables sensory dataset synthesis from existing RGB robotic datasets without dedicated sensor data generators.}

Given the RGB observation $\boldsymbol{o}_{\mathrm{RGB}}$ and spatially aligned 2D sensor observation $\boldsymbol{s}$, a segmentation module $f_{seg}$ produces a binary mask $\boldsymbol{M}$ highlighting the target region specified by $\boldsymbol{l_d}$, and the grounded sensor image is constructed as:
\begin{equation}
\boldsymbol{M} = f_{seg}(\boldsymbol{o}_{\mathrm{RGB}}, \boldsymbol{l_d}), \quad
\boldsymbol{m} = \boldsymbol{M} \odot \boldsymbol{s} + (1-\boldsymbol{M}) \odot \boldsymbol{o}_{\mathrm{RGB}}
\end{equation}
We leave sensor data processing and alignment details in \Cref{appendix:impl-details}.


%% file: sections/method.tex
\section{Methodology}
\label{sec:method}

\begin{figure*}
    \centering
    \begin{minipage}[t]{0.4\textwidth}
        \centering
        \includegraphics[width=\linewidth]{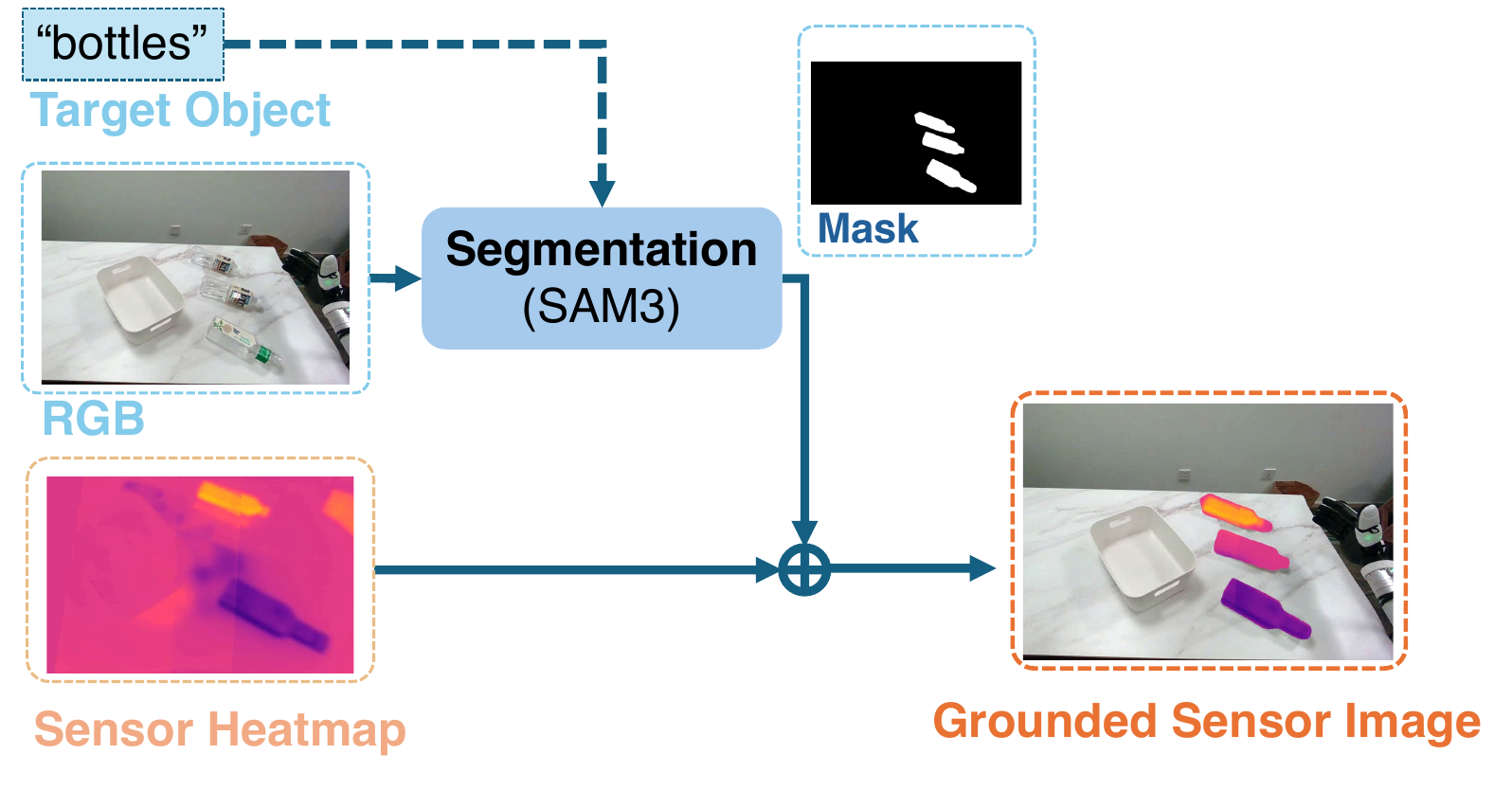}
        \caption{\textbf{Grounded sensor image processing.} We perform semantic segmentation with target description and overlay sensor heatmap at masked RGB regions.} 
        \label{fig:sensor-masked}
    \end{minipage}
    \hfill
    \begin{minipage}[t]{0.55\textwidth}
        \centering
        \includegraphics[width=\linewidth]{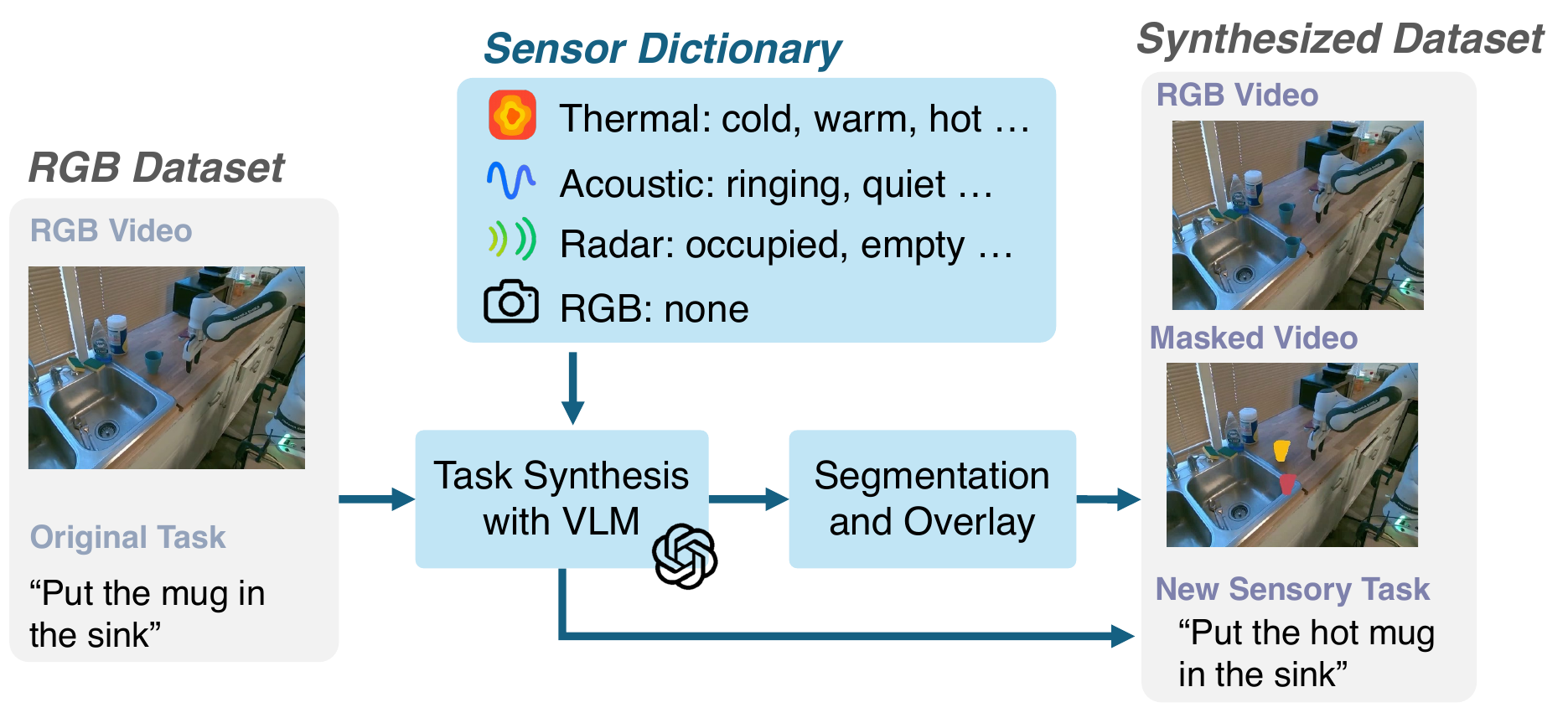}
        \caption{\textbf{Data synthesis pipeline.}
        We inject physical keywords into task instructions of existing RGB datasets and overlay color-coded masks on segmented target objects to synthesize multisensory training data.}
        \label{fig:data-synthesis}
    \end{minipage}
\end{figure*}


\subsection{\name{} Model}

\heading{Model architecture.}
As shown in \Cref{fig:model-overview},
\name{} uses pretrained VLM, PaliGemma-2~\citep{steiner2024paligemma}, as backbone for its transferable vision-language understanding and generation capability. 
The VLM consists of a Gemma-2~\citep{team2024gemma} language model and a SigLIP~\citep{zhai2023siglip} vision encoder with linear projection for alignment.
It takes in one or a sequence of images and a language instruction.
For the action expert, we use a diffusion transformer (DiT) model~\citep{peebles2023DiT}.
We use the pretrained VLM weights from VITRA~\citep{li2025vitra} as initialization to utilize its large-scale human-hand data based pretraining.

\heading{Sensor tokens.}
To enable adaptive sensor selection, we introduce learnable \textit{sensor tokens} representing different sensing modalities:
(i) \textbf{\texttt{<None>}} for standard RGB input. This allows the model to fall back to RGB when no sensing is needed.
(ii) \textbf{\texttt{<Thermal>}} for thermal input, useful for temperature-sensitive manipulation tasks. 
(iii) \textbf{\texttt{<Acoustic>}} for acoustic input. This enables tasks requiring sound awareness, such as locating objects based on sound cues, monitoring environmental sounds, etc.
(iv) \textbf{\texttt{<mmWave>}} for mmWave radar input. This allows tasks needing radar perception, such as detecting hidden objects or visually occluded objects.
These tokens serve as representative examples, while our approach readily scales to more modalities.
\name{} generates the sensor token based on the task instruction and RGB image, similar to other discrete language tokens.

\heading{Target description and segmentation.}
As shown in \Cref{fig:sensor-masked}, we construct grounded sensor images by overlaying sensor heatmaps on masked RGB regions.
The VLM backbone generates a target description $\boldsymbol{l_d}$ specifying the objects of interest for \majorEdit{robotic manipulation, together with the sensor token $\boldsymbol{l_s}$}, treating this as a language grounding \majorEdit{(image captioning)} task based on the task instruction and RGB observation (\Cref{equ:vla-process}).
For example, given ``Pick up the hot mug'', the model generates ``the mugs'' as the target description.
We then use a pretrained segmentation model, SAM3~\citep{carion2025sam3}, to obtain a binary mask of the described objects and overlay the sensor heatmap, producing grounded sensor images that encode both visual and physical sensing information in a unified RGB space.
\minorEdit{We choose SAM3 for its strong generalization to unseen objects. Keeping segmentation separate from the VLA backbone also enables asynchronous mask updates during action generation, reducing latency overhead.}


\heading{Sensor-guided action generation.}
After obtaining the grounded sensor image $\boldsymbol{m}$, we append it to the model input after encoding with the VLM vision encoder.
Due to the unified RGB space representation, we reuse the same vision encoder for all sensing modalities without requiring specialized encoders.
We append a learnable cognition token as extra input as in VITRA.
The model then generates a conditioning feature for action expert based on the task instruction, RGB observation, \majorEdit{generated} sensor token, target description, and grounded sensor image, as in \Cref{fig:model-overview}.
The action expert takes in the \majorEdit{visual-language} feature, robot states, and noisy actions to predict the denoised action sequence for robotic manipulation.

\subsection{Data Synthesis Pipeline}
Compared with RGB video based robotic datasets, 
multisensory datasets are very scarce.
Moreover, raw sensor data have distinct hardware-dependent data formats and distributions, making it difficult to simulate or synthesize realistic sensor data.
To address this data scarcity issue, we develop a data synthesis pipeline to generate multisensory datasets from existing RGB-only robotic datasets.
The key insight is that our grounded sensor image representation encodes only color-coded physical information for target objects in the RGB space, making it amenable to data synthesis. 
\Cref{fig:data-synthesis} shows the data synthesis pipeline.

\heading{Data synthesis.}
We construct a sensor dictionary mapping each modality to a diverse set of physical property keywords (e.g., thermal $\rightarrow$ ``hot'', ``cold'', ``warm'').
For each RGB video episode, we randomly sample a modality and keyword, inject the keyword into the task instruction (e.g., ``Pick up the \textit{hot} mug''), and prompt a VLM (GPT-5.2) to generate a target description. SAM3 then segments the described object, and we overlay a color corresponding to the keyword on the masked region to simulate sensor observations (\Cref{fig:data-synthesis}).
For example, for the ``hot'' keyword, we overlay a yellow color mask on the segmented object region.
The color mapping is randomized around a base color per keyword for diversity.
Lastly, we combine the color-masked video with the original RGB video as synthesized data.
We also apply a data augmentation that clones the masked region to a shifted location with a wrong sensor color, encouraging the model to rely on sensor information rather than spatial priors.

\heading{Dataset details.}
We apply the data synthesis pipeline on \majorEdit{three} existing RGB video robotic manipulation datasets, including \majorEdit{MolmoAct~\citep{molmoact2025}, AgiBotWorld-Alpha~\citep{contributors2024agibotworldrepo}}, and VITRA~\citep{li2025vitra} to build a multisensory dataset that covers diverse scenarios, tasks, and manipulation actions. \majorEdit{In total, we synthesize 9.6K episodes with 1.05~M frames, covering over 1000 objects. We randomly synthesize different sensing modalities on each episode to further increase the diversity of the synthesized dataset.}

\subsection{Training with Multisensory Data}
\name{} training includes two parts (\Cref{fig:model-overview}): VLM training for adaptive sensor selection and target description generation, and VLA training for sensor-guided action generation. 
We mix real-world collected multisensory dataset on our downstream robotic tasks together with the synthesized dataset mentioned above, which enables large-scale pretraining for zero-shot generalization on unseen tasks in \Cref{sec:eval-pretraining}.

\heading{VLM training loss.}
Using the task instruction and RGB video as input, we train the VLM backbone to generate the appropriate sensor token and target description.
We optimize the following VLM loss:
\begin{equation}
\mathcal{L}_{\text {VLM}} = \mathcal{L}_{\text {sensor}} + \mathcal{L}_{\text {target}}
\end{equation}
where $\mathcal{L}_{\text {sensor}}$ and $\mathcal{L}_{\text {target}}$ are the cross-entropy losses for sensor token and target description generation.
\majorEdit{}

\heading{VLM and VLA co-training.}
For VLA training, we incorporate the grounded sensor images as additional input and jointly train the VLM backbone with the diffusion action expert end-to-end.
The overall training loss combines the VLM loss with the diffusion MSE loss~\citep{peebles2023DiT}:
\begin{equation}
\mathcal{L} = \mathcal{L}_{\text{VLM}} + \lambda \, \mathbb{E}_{\tau, \epsilon}\!\left[\left\|\epsilon - \epsilon_{\theta}\!\left(\mathbf{a}^{\tau}, \tau, \mathbf{c}\right)\right\|_2^2\right],
\end{equation}
where $\mathbf{a}^{\tau}$ is the noisy action at diffusion step $\tau$, $\mathbf{c}$ denotes the conditioning inputs (visual-language feature from VLM and robot states), $\epsilon$ and $\epsilon_{\theta}$ are the ground-truth and predicted noise.
$\lambda$ is a hyperparameter balancing the two objectives and we set $\lambda=1\mathrm{e}{-2}$ in our experiments. 
Our joint optimization ensures the model maintains both accurate VLM outputs alongside effective action generation. This is critical as training VLA alone without the VLM loss leads to degraded VLM outputs according to our experiments. 



%% file: sections/eval.tex
\begin{figure}
    \centering
    \begin{subfigure}[t]{0.55\columnwidth}
        \centering
        \includegraphics[width=\linewidth]{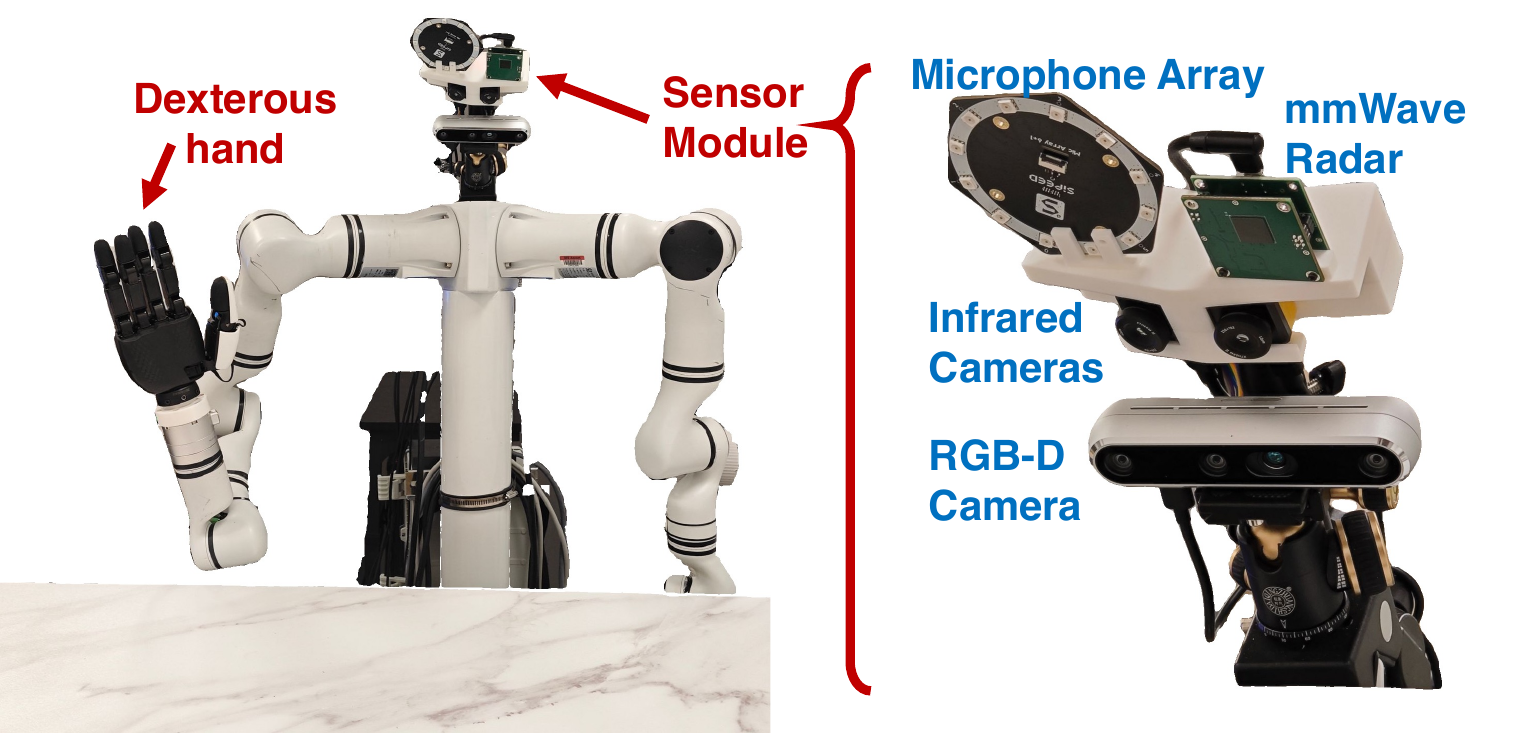}
        \caption{Robot hardware setup.}
        \label{fig:eval-hardware-setup-subfig}
    \end{subfigure}\hfill
    \begin{subfigure}[t]{0.43\columnwidth}
        \centering
        \includegraphics[width=\linewidth]{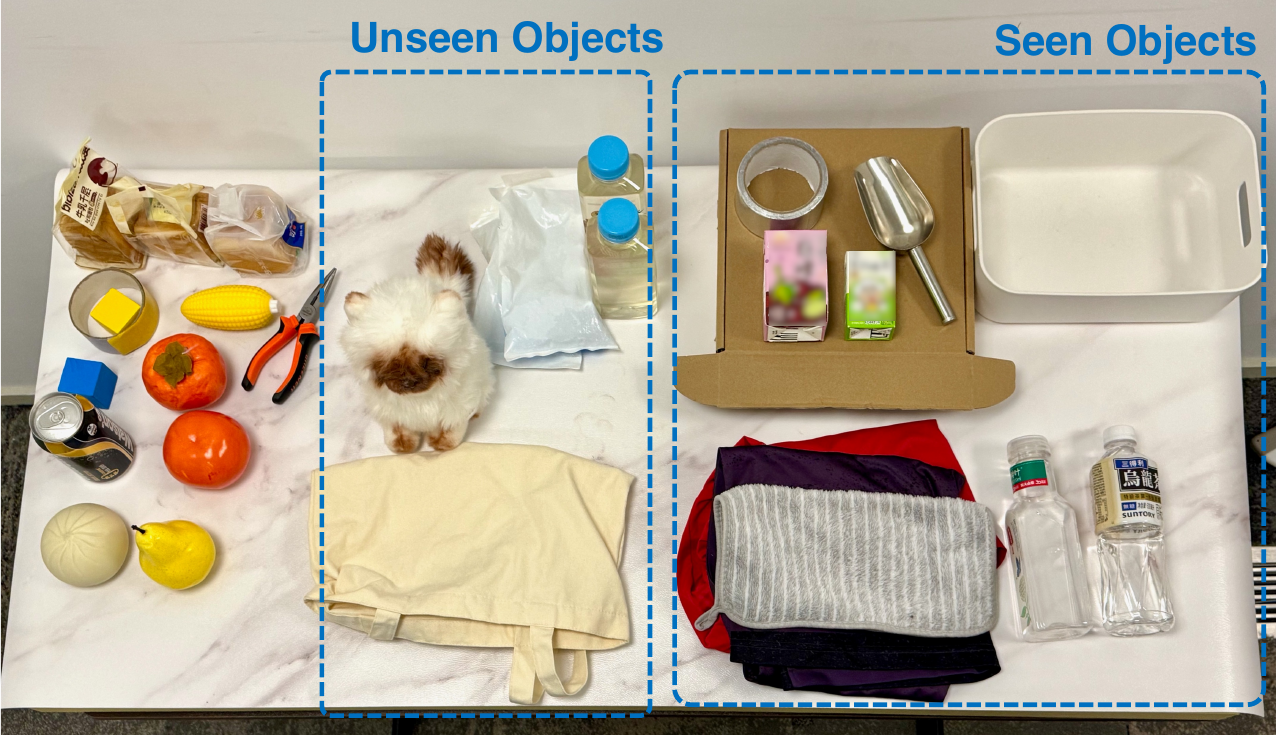}
        \caption{Objects involved in the tasks.}
        \label{fig:eval-task-objects}
    \end{subfigure}
    \caption{Evaluation setup. We set up a robot arm with a 12DoF dexterous hand and a multi-sensor module and evaluate \name{} on a variety of task objects spanning different sensory modalities.}
    \label{fig:eval-hardware-setup}
\end{figure}

\section{Experiments}
\label{sec:eval}
Our evaluation covers the following:
(1) task success rates on a set of challenging manipulation tasks that require multimodal sensing inputs.
(2) adaptive sensor selection performance and improved efficiency. 
(3) benefits of synthesized dataset pretraining, showing generalization to unseen tasks.

\begin{table*}[t]
\small
\centering
\setlength{\belowcaptionskip}{0pt}
\setlength{\tabcolsep}{4pt}
\caption{Task success rates comparison across three types of sensing modality tasks.} 
\begin{tabular}{lccccS[round-mode=figures, round-precision=3, table-format=2.1, table-space-text-post={\%}] S[round-mode=figures, round-precision=3, table-format=2.1, table-space-text-post={\%}] S[round-mode=figures, round-precision=3, table-format=1.3]c}
\toprule
 & \multicolumn{4}{c}{Per-modality Success Rate} & \multicolumn{3}{c}{Per-stage Success Rate} & \\
\cmidrule(lr){2-5}\cmidrule(lr){6-8}
 & Thermal & Acoustic & mmWave & \textit{Average} & {Sensing} & {Manipulation} & {\textit{Score}} & $p$ \\
\midrule
\piZero-RGB  & 33.3\% & 25.0\%  & 4.17\% & 20.8\% & 48.6\% & 43.1\% & 0.458 & $<$0.001 \\
\piZeroFive-RGB  & 16.7\% & 33.3\%  & 8.33\% & 19.4\% & 41.7\% & 33.3\% & 0.375 & $<$0.001 \\
\piZero-Raw  & 16.7\% & 41.7\% & 25.0\% & 27.8\% & 86.1\% & 27.8\% & 0.569 & $<$0.001 \\
\piZeroFive-Raw  & 16.7\% & 33.3\%  & 20.8\% & 23.6\% & 83.3\% & 29.2\% & 0.563 & $<$0.001 \\
\midrule
\name{}-RGB  & 12.5\% & 33.3\%  & 20.8\% & 22.2\% & 41.7\% & 43.1\% & 0.424 & $<$0.001 \\
\name{}-Raw  & 41.7\% & 25.0\% & 33.3\% & 33.3\% & 91.7\% & 33.3\% & 0.625 & $<$0.001 \\
\name{}-RawAdapt  & 70.8\% & 41.7\% & 66.7\% & 59.7\% & 93.1\% & 59.7\% & 0.764 & $<$0.05 \\
\textbf{\name{} (Ours)} & \textbf{83.3\%} & \textbf{58.3\%} & \textbf{87.5\%} & \textbf{76.4\%} & \textbf{95.8\%} & \textbf{77.8\%} & \textbf{0.868} & -- \\
\bottomrule
\end{tabular}
\label{tab:main-result}
\end{table*}


\begin{table}[t]
\small
\centering
\setlength{\tabcolsep}{6pt}
\setlength{\belowcaptionskip}{0pt}
\caption{Adaptive sensor selection accuracy.}
\begin{tabular}{lcccc}
\toprule
 & \multicolumn{2}{c}{Training Tasks} & \multicolumn{2}{c}{Unseen Tasks} \\
\cmidrule(lr){2-3}\cmidrule(lr){4-5}
 & Sensor & Target & Sensor & Target \\
\midrule
PaliGemma-2         & 0\% & 13.0\% & 0\% & 9.5\% \\
\name{} w/o pretrain    & {100\%} & \textbf{100\%} & {85\%} & {40.5\%} \\
\textbf{\name{} (pretrained)}    & \textbf{100\%} & {93.5\%} & \textbf{100\%} & \textbf{82.0\%} \\
\bottomrule
\end{tabular}
\label{tab:sensor-selection}
\end{table}

\begin{table*}[t]
\footnotesize
\centering
\setlength{\tabcolsep}{3.5pt}
\setlength{\belowcaptionskip}{0pt}
\caption{Impact of pretraining on synthesized multisensory data. We report success rates on training tasks and zero-shot generalization to unseen tasks.}
\begin{tabular}{lccccccccc}
\toprule
 & \multicolumn{4}{c}{Seen Tasks} & \multicolumn{5}{c}{Unseen Tasks} \\
\cmidrule(lr){2-5}\cmidrule(lr){6-10}
 & Thermal & Acoustic & mmWave & \textit{Average} & Thermal & Acoustic & mmWave & \textit{Average} & $p$ \\
\midrule
\name{}-Raw  & 41.7\% & 25.0\% & 33.3\% & 33.3\% & 31.3\% & 25.0\% & 18.8\% & 25.0\% & $<$0.001 \\
\name{} w/o pretrain  & 83.3\% & 58.3\% & \textbf{87.5\%} & 76.4\% & 25.0\% & 31.3\% & 25.0\% & 27.1\% & $<$0.001 \\
\textbf{\name{} (pretrained)} & \textbf{87.5\%} & \textbf{70.8\%} & 83.3\% & \textbf{80.6\%} & \textbf{75.0\%} & \textbf{56.3\%} & \textbf{68.8\%} & \textbf{66.7\%} & -- \\
\bottomrule
\end{tabular}
\label{tab:pretrain-compare}
\end{table*}

\subsection{Experiment Setup}

\heading{Robot setup.}
We set up a table-top experiment environment using a robot equipped with 12-DoF Robotera XHand dexterous hands and a sensor suite, as shown in \Cref{fig:eval-hardware-setup}.
The sensor suite includes an RGB-D camera (Intel RealSense), two thermal cameras (infiRay T2S), an mmWave radar (Calterah 4T4R 60GHz radar), and a microphone array (Sipeed 6+1Mic Array).
Each sensor provides a 2D heatmap image aligned with the RGB camera's field of view, e.g., the thermal camera provides a temperature heatmap, the mmWave radar provides a reflection intensity heatmap, and the microphone array provides a sound intensity heatmap. (\Cref{appendix:impl-details}.)
To collect robotic manipulation demonstration data, we use a teleoperation system with a pair of MANUS5 gloves to control the robot hands and record the action sequences together with all sensors' measurements.


\heading{Task setting.}
We design three categories of manipulation tasks that demand beyond-RGB sensing modalities, each with multiple sub-task instruction variants and a set of target objects.
(1) \textit{Thermal-guided pick-and-place}: The robot picks up a drink with a specified temperature and places it into a basket, e.g., \textit{``Pick up the hot drink and place it into the basket.''} Instructions cover three thermal keywords (\textit{hot}, \textit{cold}, \textit{room-temperature}) over two types of drinks with varying positions and orientations.
(2) \textit{Acoustic-grounded object search}: The robot localizes a hidden sound source using spatial audio cues and removes the covering, e.g., \textit{``Pick up the clothes/towels covering the ringing phone and place it into the basket''}. Coverings include clothes and towels of varying shapes and colors.
(3) \textit{mmWave radar-guided object search}: The robot uses the mmWave radar to see through closed boxes, opens the box that contains an item, e.g., \textit{``Open the occupied box.''} Hidden items include drinks, shovels, and tape rolls, with box positions varied across trials.
In total, we collect 720 teleoperated demonstration episodes, covering 10 different sub-task instructions, 7 objects with varying placements across 3 sensing modalities, as our real-world training dataset.
\Cref{fig:eval-task-objects} shows seen objects in our datasets. \Cref{fig:intro-overview} shows task examples and \Cref{fig:frames-seen-tasks} shows examples of successful action trajectories. We evaluate tasks with various distracting objects in the scene to test the model's ability to ground on the correct target object.

\heading{Training setup.}
We initialize from VITRA~\cite{li2025vitra} weights for both VLM backbone and action expert, and perform VLM and VLA co-training on 64 A100 (40GB) GPUs with a batch size of 512 for 20K steps (${\sim}$20 hours) using a learning rate of $1\mathrm{e}{-5}$ and AdamW optimizer.
We train and evaluate a single \name{} model across all tasks, which is more challenging than per-task models.

\heading{Evaluation metrics.}
We report two types of metrics. (i) \textit{Task success rate} is the fraction of trials in which the robot completes the full task instruction end-to-end. (ii) \textit{Task score} decomposes a trial into a \emph{sensing} sub-task and a \emph{manipulation} sub-task, each contributing $0.5$: $0.5$ for selecting and grounding on the correct target object (e.g., the hot drink, the occupied box), and $0.5$ for performing the correct manipulation (picking and placing in the container, opening the box, removing the covering). The per-stage \textit{sensing} and \textit{manipulation} success rates in \Cref{tab:main-result} are the average across all task trials.
\minorEdit{We also report Fisher's exact two-sided $p$-values of the pooled-average gap against \name{}.}

\heading{Baselines.} 
We compare against two baseline types by finetuning frontier VLA models, \piZero{} and \piZeroFive{}, on our dataset:
(i) \textit{VLA-RGB}, standard RGB-only VLA models that receive no sensor input;
(ii) \textit{VLA-Raw}, VLA models that receive spatially aligned sensor heatmaps from all available sensors concatenated as additional image inputs, but without the grounded sensor image processing or adaptive sensor selection.


\subsection{Multisensory Manipulation Task Performance}

\heading{Task success rates.}
We show the task success rates and task scores of \name{} and baselines in \Cref{tab:main-result}, with each per-modality cell averaged over $24$ independent trials. 
\name{} (Ours) here is trained on the real-world dataset only; the additional benefit of pretraining on synthesized data is reported in \Cref{tab:pretrain-compare}.
We achieve high success rates across all three sensing modalities, with acoustic tasks lower due to challenges of grasping soft fabric/clothes.
Our approach achieves an average success rate of \minorEdit{76.4\%} across all tasks, significantly outperforming the RGB-only VLA baselines (\minorEdit{22.2\%} for \name{}-RGB and \minorEdit{20.8\%} for \piZero-RGB), demonstrating the necessity of multimodal sensing input.
For VLA baselines with raw sensor heatmaps as input, the performance is \minorEdit{33.3\%} of \name{}-Raw and \minorEdit{27.8\%} of \piZero-Raw, which is substantially lower than our approach.
We observe that they can select the correct target object in the sensing stage, but fail to perform the correct manipulation, indicating that the raw sensor heatmaps are noisy and not suited for manipulation.
This shows that naively inputting all sensor heatmaps without proper processing and adaptive selection limits the benefits of multisensory perception.



\heading{Ablation study.}
To validate the effectiveness of key components in \name{}, we conduct ablation studies by (1) removing the grounded sensor image representation (\name{}-RawAdapt) and (2) removing both grounded sensor image and the adaptive sensor token prediction mechanism (\name{}-Raw). From \Cref{tab:main-result}, we observe that removing either component leads to performance drops, demonstrating the importance of both grounded sensor images for unified sensor fusion and adaptive sensor selection for efficient multisensory perception. 
We also note that \name{}-RawAdapt outperforms \name{}-Raw significantly, indicating that adaptively selecting the most relevant sensor heatmap boosts the task success rate, by reducing the input noise from irrelevant sensors and allowing the model to focus on the most informative sensor modality for each task.

\heading{Multi-stage multi-sensor tasks.}
To showcase \name{}'s ability to invoke different sensors within long-horizon tasks, we design two multi-stage tasks that chain heterogeneous sensing and manipulation stages. 
(i) \textit{Radar~$\rightarrow$~RGB}: the robot uses mmWave radar to identify which closed box contains an item, opens it, then performs RGB-only pick-and-place into a basket. (ii) \textit{Radar~$\rightarrow$~Thermal}: after radar-guided box opening, the robot picks a drink of a specified temperature (hot, cold, or room-temperature) from multiple drinks, requiring a switch from radar to thermal sensing. Example trajectories are shown in \Cref{fig:frames-multi-stage-tasks}.
Over $12$ trials per task, Radar~$\rightarrow$~RGB achieves 66.7\% end-to-end success (100\% box localization, 83.3\% opening, 66.7\% pick-and-place), and Radar~$\rightarrow$~Thermal achieves 75.0\% (100\% localization and thermal selection, 100\% opening, 75.0\% pick-and-place). These results confirm that \name{} can adaptively dispatch the appropriate sensor at each stage and chain heterogeneous modalities within a single end-to-end model.







\subsection{Adaptive Sensor Selection}

\heading{VLM generation accuracy.}
We evaluate the adaptive sensor selection of \name{} in \Cref{tab:sensor-selection} across 200 samples each from training and unseen tasks.
We measure two accuracy metrics:
 (1) \textit{sensor selection accuracy}, the fraction of correct sensor token predictions, and (2) \textit{target description accuracy}, the fraction of semantically correct target descriptions (evaluated via GPT-5.2).
\name{} substantially outperforms the PaliGemma-2 base VLM on both metrics, demonstrating the effectiveness of our VLM training.
Pretraining on synthesized data further boosts target description accuracy from 40.5\% to 82.0\% on unseen tasks, which is critical for zero-shot generalization.

\heading{Inference efficiency.}
By selecting only the task-relevant sensor, \name{} reduces inference GPU memory from 13.23~GB to 6.61~GB compared to \name{}-Raw (excluding 11.9~GB model weights, averaged over 100 samples). This efficiency gain applies to both training and inference, as the model can train with larger batch sizes. 
More importantly, our memory cost does not increase with the number of sensors, unlike the rapid growth if feeding all sensor inputs.


\subsection{Effectiveness of Pretraining}
\label{sec:eval-pretraining}

\begin{figure*}[t]
    \centering
    \setlength{\belowcaptionskip}{0pt}
    \includegraphics[width=1\linewidth]{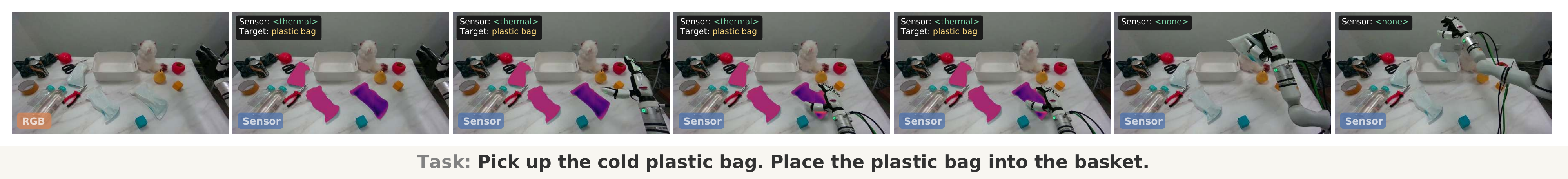}
    \includegraphics[width=1\linewidth]{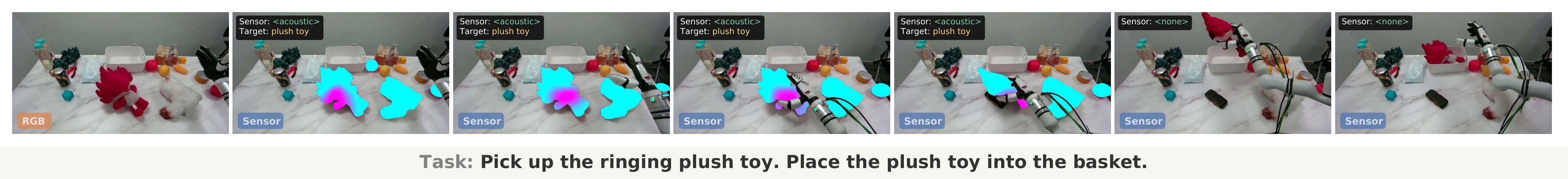}
    \includegraphics[width=1\linewidth]{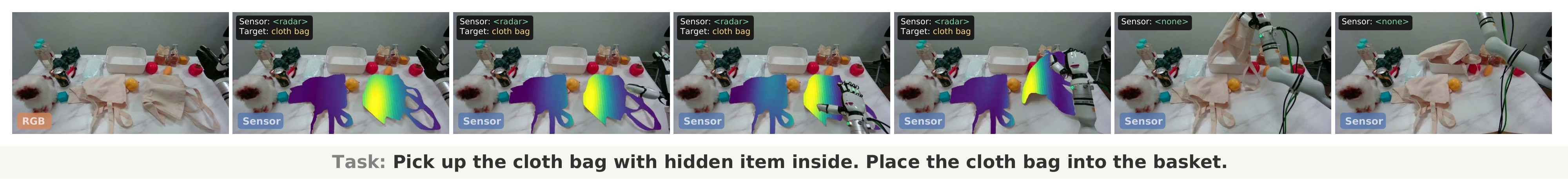}
    \caption{Execution trajectories of unseen task evaluation.}
    \label{fig:eval-unseen-tasks}
\end{figure*}

\heading{Unseen task setting.}
To probe the generalization of \name{}, we evaluate on a suite of \emph{unseen tasks} that never appear in training, with $16$ independent trials per task. These instructions recombine sensing modalities with new manipulation targets, e.g., \textit{``Pick up the cloth bag with the ringing item inside and place it into the basket.''}, which couples acoustic localization with grasping a previously unseen object. The model must understand the new instruction, invoke the correct sensor, and ground it on the unfamiliar object to complete the task. \Cref{fig:eval-unseen-tasks} shows \name{} robot execution trajectories on these unseen tasks, with more details in \Cref{fig:frames-unseen-tasks}. 

\heading{Pretraining effectiveness.}
We compare \name{} with and without pretraining on synthesized data in \Cref{tab:pretrain-compare}, using $24$ trials per modality on seen tasks and $16$ trials per modality on unseen tasks.
\name{} with pretraining shows comparable performance on seen tasks, while significantly improving unseen task success by 39\% on average (from 27.1\% to 66.7\%), with gains across all three modalities.
We observe a gain of \minorEdit{12\%} for seen acoustic tasks due to improved dexterous hand manipulation when grabbing the soft clothes.
The gain for unseen zero-shot performance stems from both more accurate sensor selection and target descriptions (\Cref{tab:sensor-selection}) and better manipulation actions learned from a more diverse set of manipulation tasks in the synthesized dataset.




%% file: sections/conclusion.tex
\section{Conclusion}
\label{sec:conclusion}
We present \name{}, an adaptive multimodal sensing VLA model that enables scalable and efficient sensor integration as on-demand tools for dexterous hand based manipulation. 
It predicts a sensor token and target description to invoke the most relevant modality, then converts the measurement into a unified \emph{grounded sensor image} for action generation. 
A data synthesis pipeline augments existing RGB videos with grounded sensor images, easing the need for large multisensory robot datasets.
On a real-world robot, \name{} reaches an 80.6\% average success rate across thermal-, audio-, and radar-guided tasks, substantially outperforms RGB-only and multisensory baselines, and generalizes to unseen tasks zero-shot, pointing toward a scalable path for general-purpose multisensory manipulation.
Current limitations include the modest scale of our real-world dataset and the dependence on a performant segmentation module, both of which can be relaxed as larger datasets and stronger segmentation models become available.


%% file: sections/appendix.tex



\section{Implementation Details}
\label{appendix:impl-details}

\subsection{Sensor Heatmap Construction}
The thermal camera directly outputs a 2D temperature map over pixel coordinates, requiring no additional spatial processing.
For the mmWave radar and microphone array, each array element $k$ records a complex-valued measurement $z_k = a_k e^{j\phi_k}$.
We convert these per-element signals into spatially resolved 2D heatmaps using standard digital beamforming, a well-established signal processing technique.
We consider the received power from a direction $(\theta, \varphi)$ (azimuth, elevation):
\begin{equation}
P(\theta,\varphi) = 20\log_{10}\!\left\lVert \sum_{k=1}^{K} a_k\, e^{j\phi_k}\, e^{-j\Delta_k(\theta,\varphi)} \right\rVert, \quad
\Delta_k(\theta,\varphi) = \frac{2\pi}{\lambda}\!\left(d_k^x \cos\varphi\,\sin\theta + d_k^y \sin\varphi\right),
\label{eq:beamforming}
\end{equation}
where $\lambda$ is the carrier wavelength and $(d_k^x, d_k^y)$ is the position of element $k$ in the array plane.
The resulting azimuth-elevation heatmap captures spatial distribution of the received power, such as mmWave reflection intensity or received sound intensity.
We map each heatmap into RGB space using sensor-specific colormaps (inferno for thermal, viridis for mmWave, plasma for acoustic), with fixed normalization ranges determined by the typical operating range of each sensor, e.g., 0--60\textdegree C for thermal.

\subsection{Sensor Heatmap Spatial Alignment}
\label{appendix:sensor-alignment}
To construct grounded sensor images, the segmentation mask produced by SAM3 in the RGB pixel space must be transferred onto each sensor's heatmap. Because the sensors are rigidly co-mounted with the RGB camera and table-top manipulation operates at a roughly constant working distance, parallax across the workspace is small relative to the heatmap angular resolution. We therefore reduce the cross-modal alignment to a one-time, per-sensor $2$D-to-$2$D projection rather than a full $3$D extrinsic calibration.

For each sensor $i$, we model the mapping from its heatmap pixel coordinates $(u_i, v_i)$ to the RGB pixel coordinates $(u, v)$ as a homography $H_i \in \mathbb{R}^{3\times 3}$:
\begin{equation}
\begin{pmatrix} u \\ v \\ 1 \end{pmatrix} \sim H_i \begin{pmatrix} u_i \\ v_i \\ 1 \end{pmatrix}.
\label{eq:sensor-homography}
\end{equation}
We estimate $H_i$ once during a one-time offline calibration procedure by manually selecting a small set of corresponding control points (well-localized landmarks visible in both the RGB image and the sensor heatmap) and solving for $H_i$ via the standard direct linear transform. No retraining or online optimization is required, and the same $H_i$ is reused across all subsequent inference.

At inference time, given an RGB-space binary mask $M$ produced by SAM3 from the target description, we project each sensor heatmap pixel into the RGB plane and look up the mask value:
\begin{equation}
M_i(u_i, v_i) = M\!\left(\pi(H_i \, [u_i, v_i, 1]^\top)\right),
\end{equation}
where $\pi(\cdot)$ denotes the perspective division. The grounded sensor image $\boldsymbol{m}$ is then formed by composing the sensor heatmap on the masked region and the RGB observation elsewhere, as in \Cref{sec:masked-images}. Because $H_i$ is fixed offline and the projection is a single matrix multiplication per pixel, the alignment introduces negligible computational overhead at runtime.





\section{Real-Robot Execution Trajectories}

\begin{figure*}[t]
    \centering
    \includegraphics[width=\linewidth]{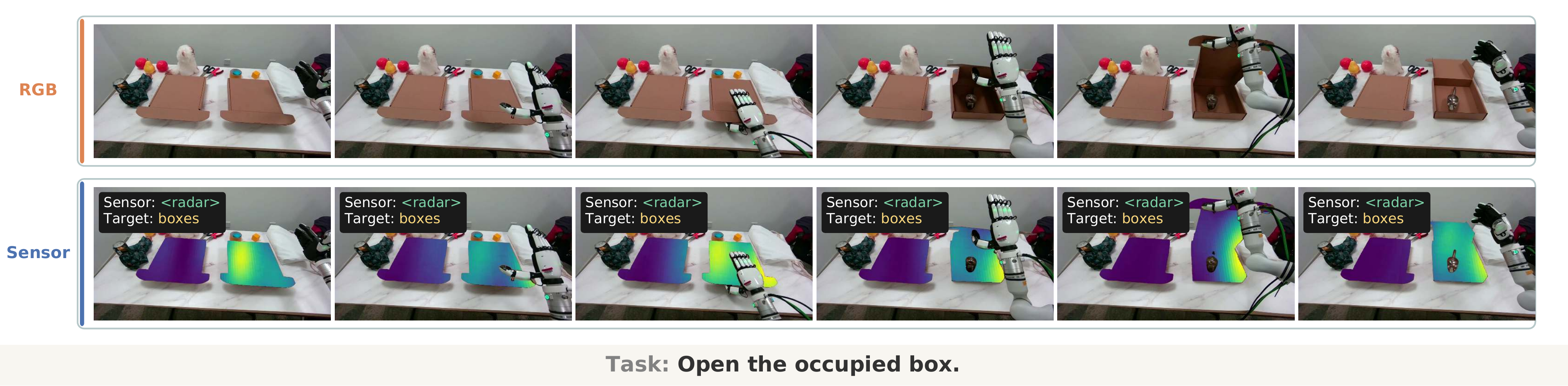}
    \includegraphics[width=\linewidth]{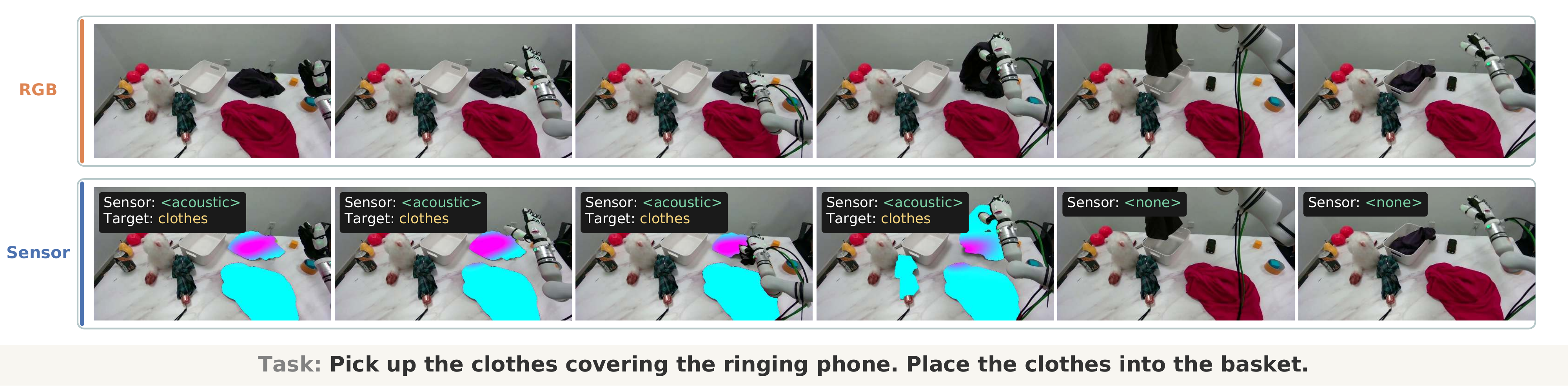}
    \includegraphics[width=\linewidth]{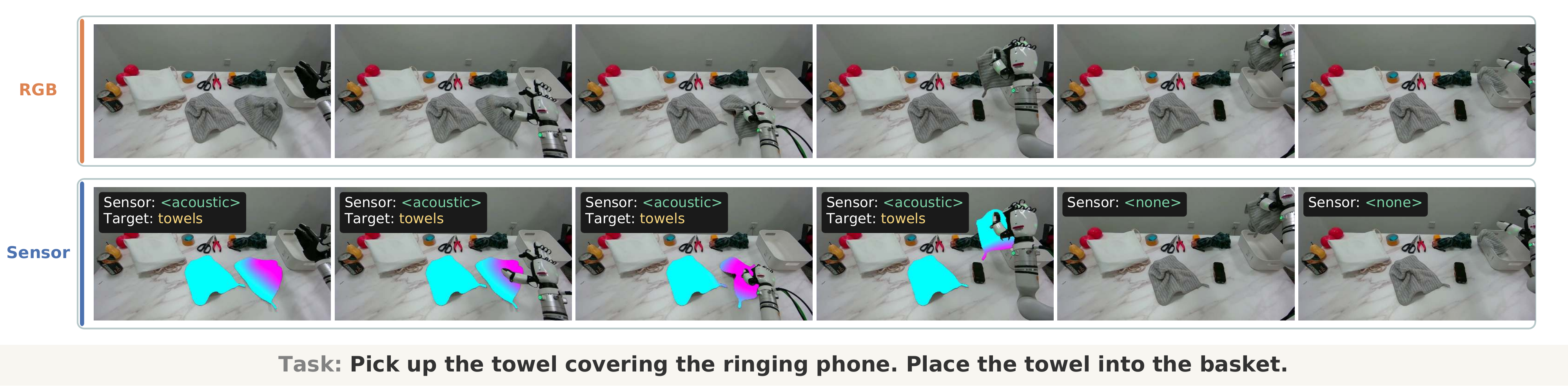}
    \includegraphics[width=\linewidth]{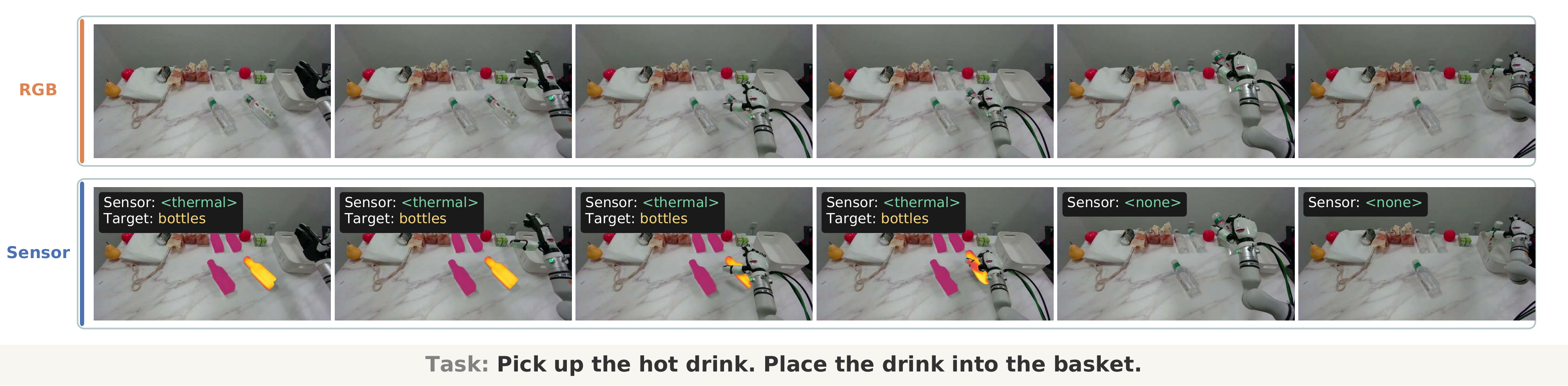}
    \includegraphics[width=\linewidth]{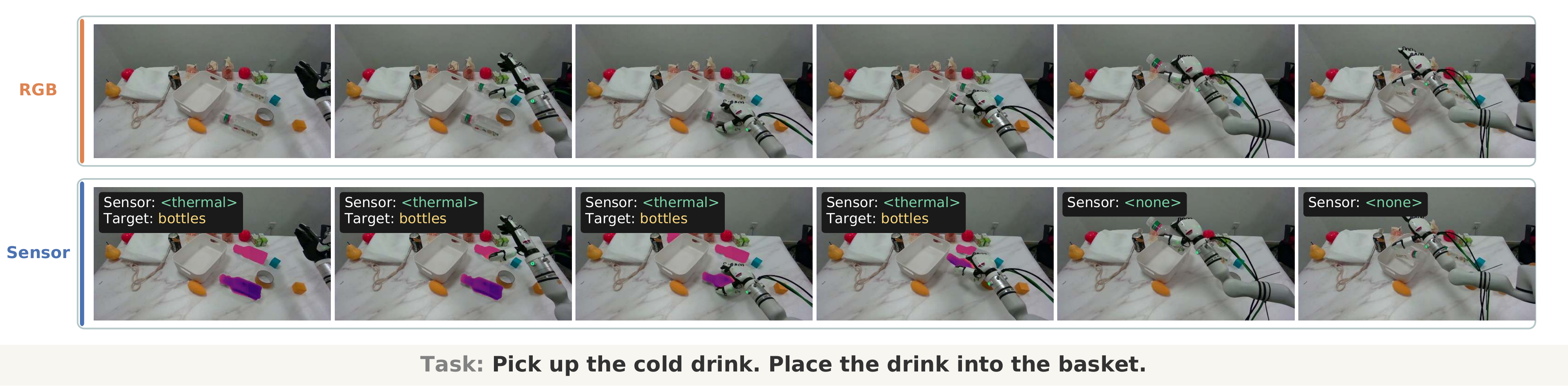}
    \caption{\name{} execution trajectories of training task evaluation.}
    \label{fig:frames-seen-tasks}
\end{figure*}

\begin{figure*}
    \centering
    \includegraphics[width=\linewidth]{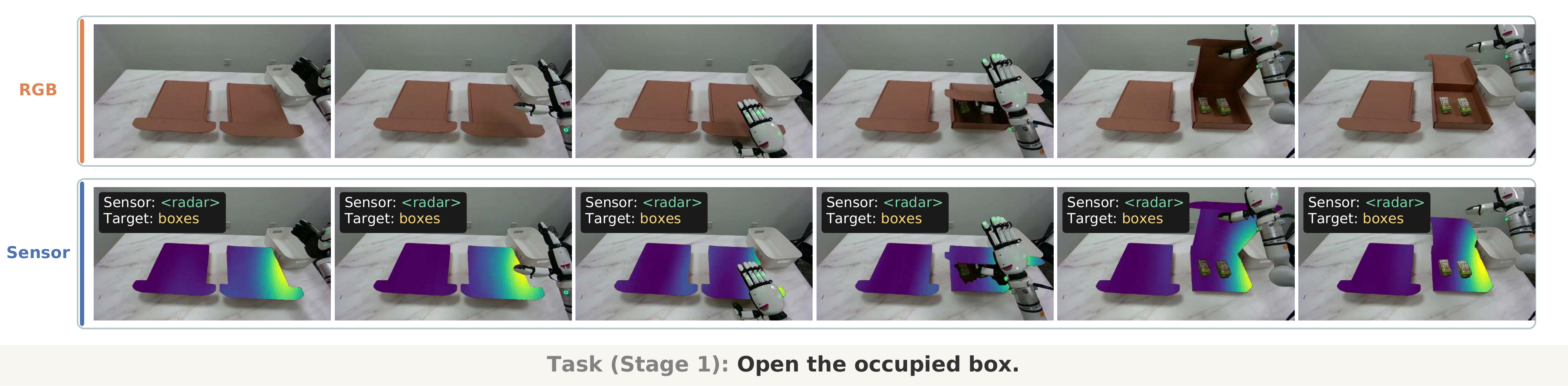}
    \includegraphics[width=\linewidth]{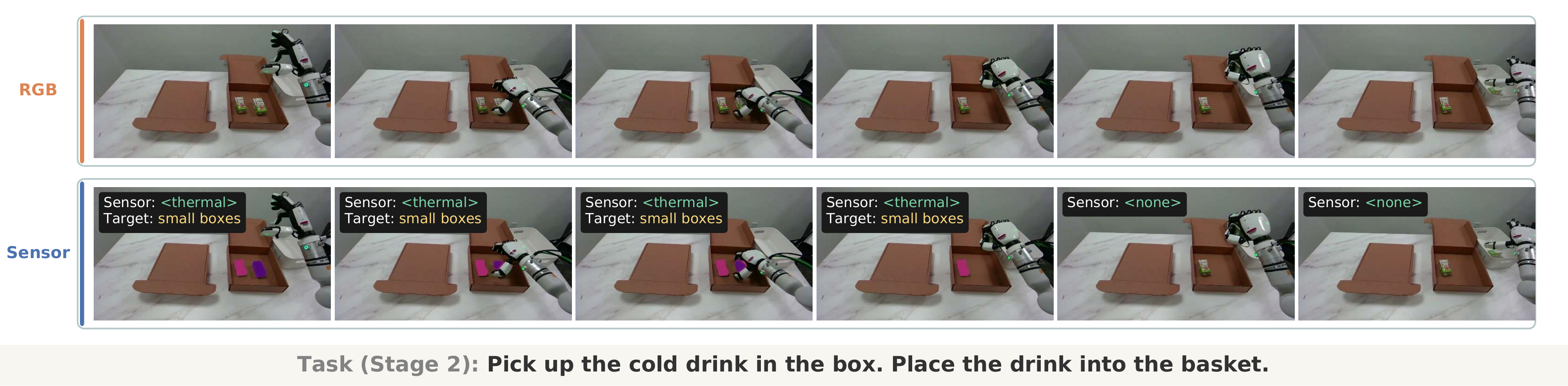}
    \includegraphics[width=\linewidth]{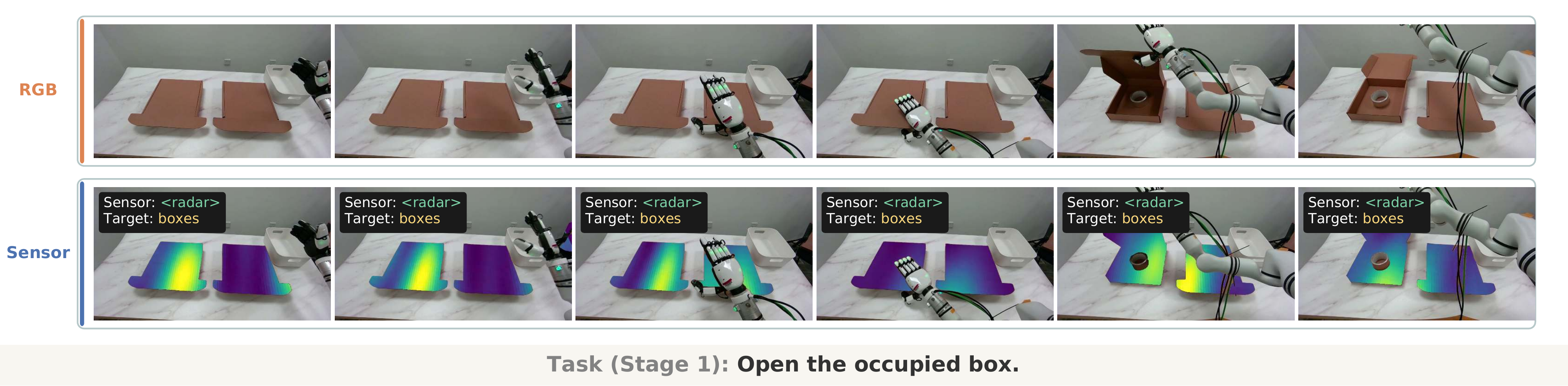}
    \includegraphics[width=\linewidth]{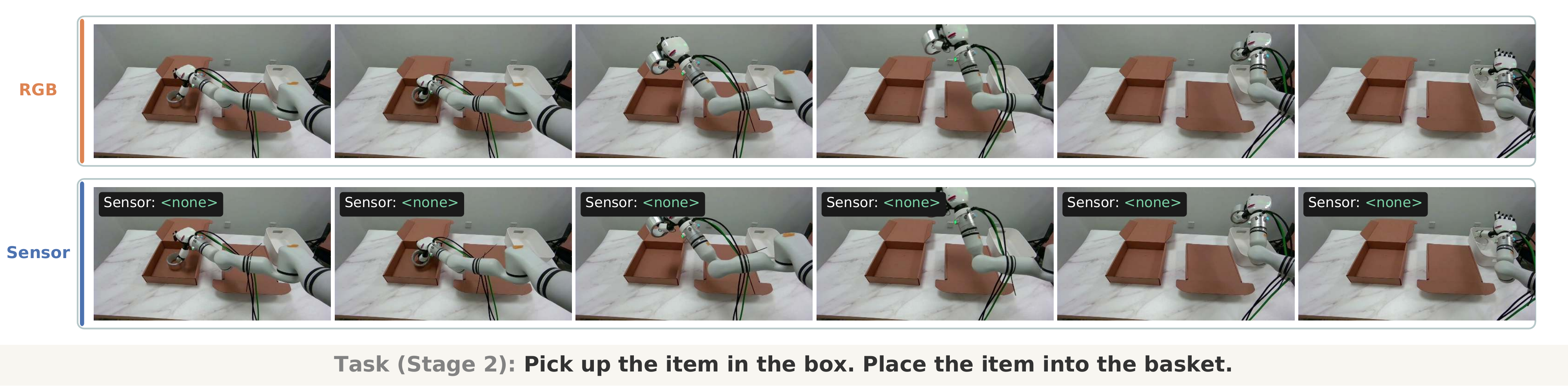}
    \caption{\name{} execution trajectories of multi-stage task evaluation.}
    \label{fig:frames-multi-stage-tasks}
\end{figure*}

\begin{figure*}
    \centering
    \includegraphics[width=\linewidth]{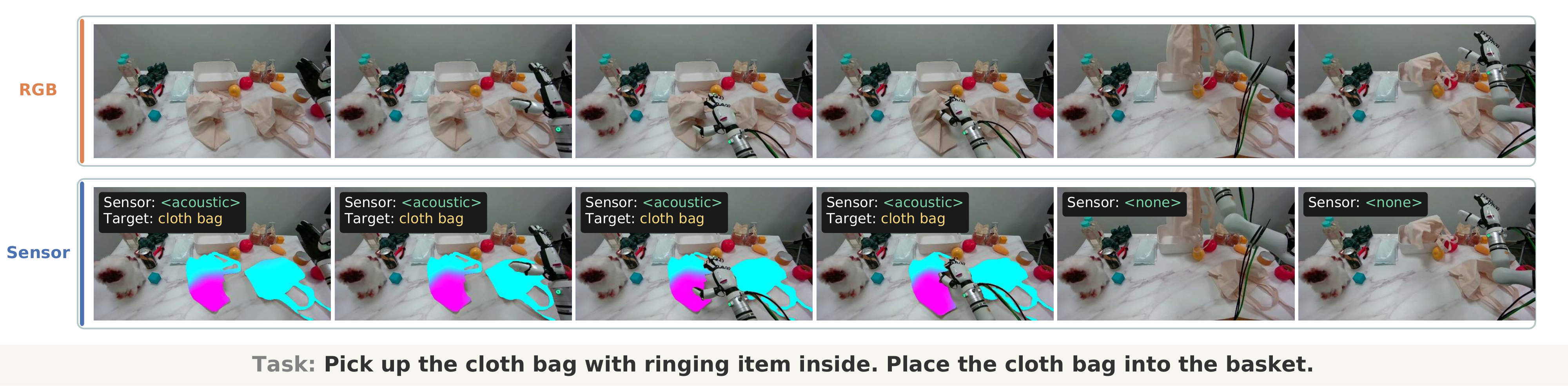}
    \includegraphics[width=\linewidth]{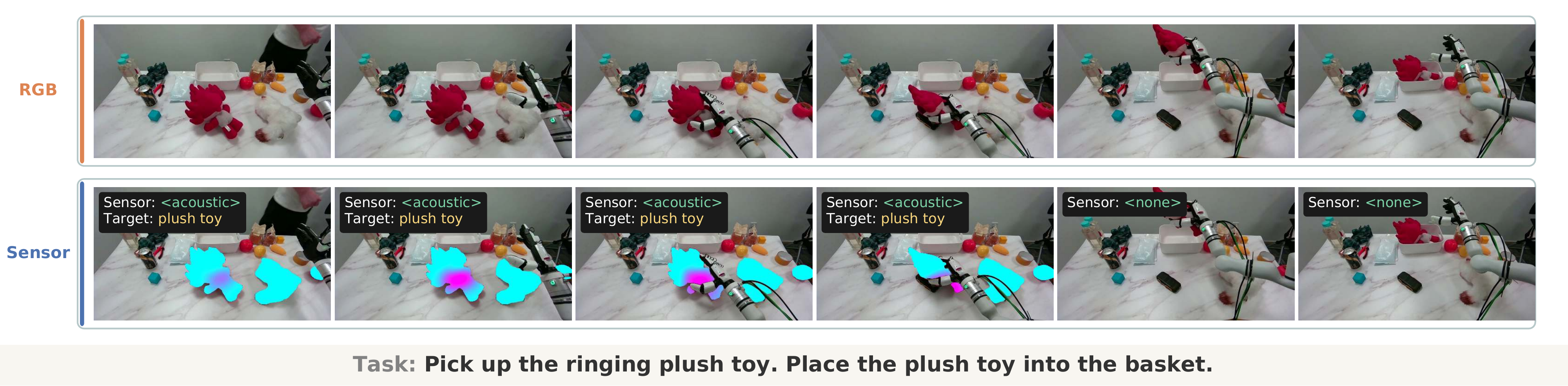}
    \includegraphics[width=\linewidth]{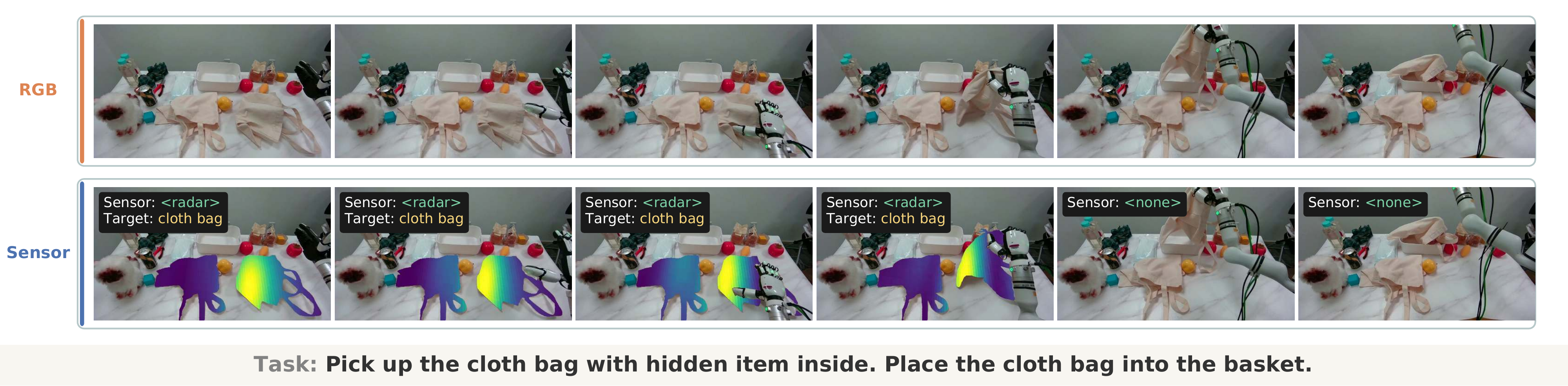}
    \includegraphics[width=\linewidth]{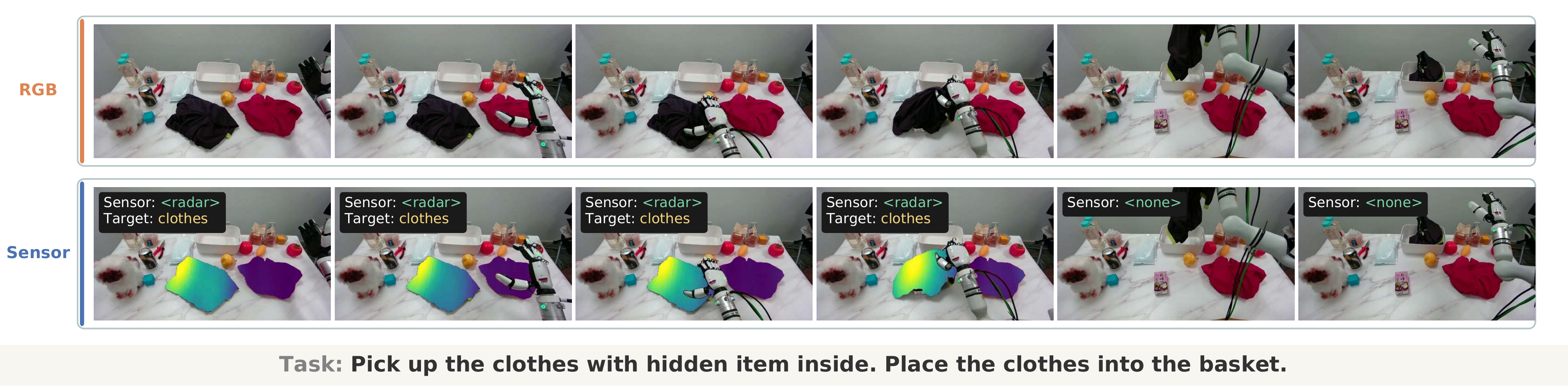}
    \includegraphics[width=\linewidth]{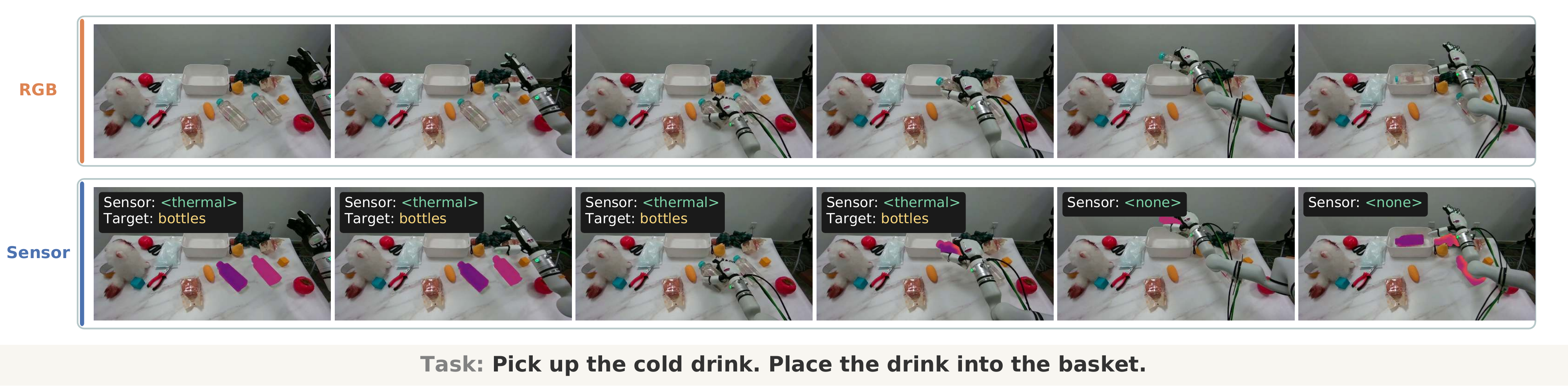}
    \includegraphics[width=\linewidth]{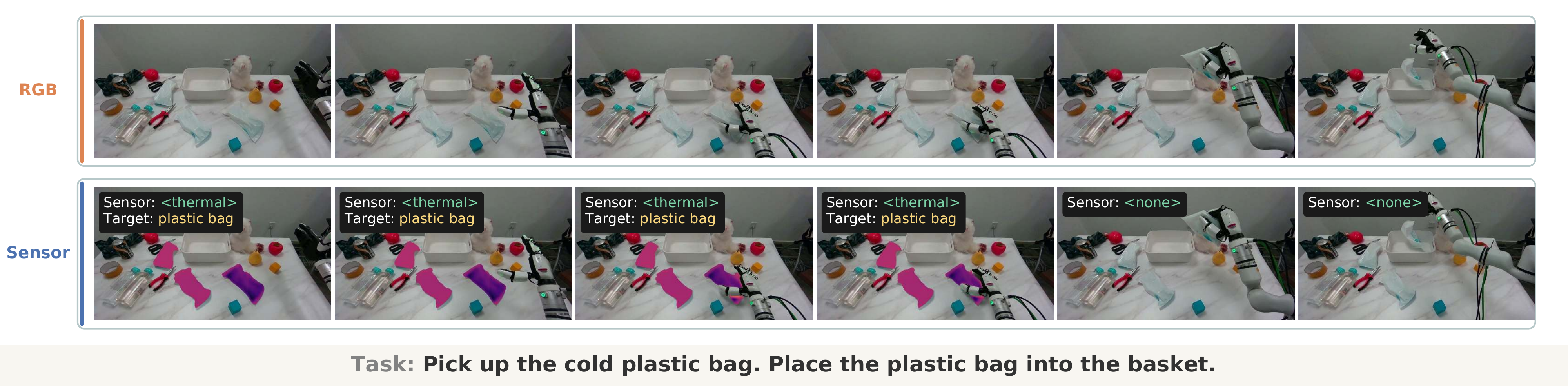}
    \caption{\name{} execution trajectories of zero-shot unseen task evaluation.}
    \label{fig:frames-unseen-tasks}
\end{figure*}




%% file: biblio.bib
@article{black2024pi0,
      title={$\pi_0$: A Vision-Language-Action Flow Model for General Robot Control}, 
      author={Kevin Black and Noah Brown and Danny Driess and Adnan Esmail and Michael Equi and Chelsea Finn and Niccolo Fusai and Lachy Groom and Karol Hausman and Brian Ichter and Szymon Jakubczak and Tim Jones and Liyiming Ke and Sergey Levine and Adrian Li-Bell and Mohith Mothukuri and Suraj Nair and Karl Pertsch and Lucy Xiaoyang Shi and James Tanner and Quan Vuong and Anna Walling and Haohuan Wang and Ury Zhilinsky},
      year={2024},
      eprint={2410.24164},
      archivePrefix={arXiv},
      primaryClass={cs.LG},
      url={https://arxiv.org/abs/2410.24164}, 
}

@article{shukor2025smolvla,
  title={{SmolVLA}: A vision-language-action model for affordable and efficient robotics},
  author={Shukor, Mustafa and Aubakirova, Dana and Capuano, Francesco and Kooijmans, Pepijn and Palma, Steven and Zouitine, Adil and Aractingi, Michel and Pascal, Caroline and Russi, Martino and Marafioti, Andres and others},
  journal={arXiv preprint arXiv:2506.01844},
  year={2025}
}

@article{wen2025dexvla,
  title={{DexVLA}: Vision-language model with plug-in diffusion expert for general robot control},
  author={Wen, Junjie and Zhu, Yichen and Li, Jinming and Tang, Zhibin and Shen, Chaomin and Feng, Feifei},
  journal={arXiv preprint arXiv:2502.05855},
  year={2025}
}

@article{ze2024generalizable,
  title={Generalizable humanoid manipulation with 3d diffusion policies},
  author={Ze, Yanjie and Chen, Zixuan and Wang, Wenhao and Chen, Tianyi and He, Xialin and Yuan, Ying and Peng, Xue Bin and Wu, Jiajun},
  journal={arXiv preprint arXiv:2410.10803},
  year={2024}
}

@inproceedings{mees2024octo,
  title={Octo: An open-source generalist robot policy},
  author={Mees, Oier and Ghosh, Dibya and Pertsch, Karl and Black, Kevin and Walke, Homer Rich and Dasari, Sudeep and Hejna, Joey and Kreiman, Tobias and Xu, Charles and Luo, Jianlan and others},
  booktitle={First Workshop on Vision-Language Models for Navigation and Manipulation at ICRA 2024},
  year={2024}
}

@article{kim2024openvla,
  title={{OpenVLA}: An open-source vision-language-action model},
  author={Kim, Moo Jin and Pertsch, Karl and Karamcheti, Siddharth and Xiao, Ted and Balakrishna, Ashwin and Nair, Suraj and Rafailov, Rafael and Foster, Ethan and Lam, Grace and Sanketi, Pannag and others},
  journal={arXiv preprint arXiv:2406.09246},
  year={2024}
}

@article{patratskiy2025spatial,
  title={Spatial Traces: Enhancing VLA Models with Spatial-Temporal Understanding},
  author={Patratskiy, Maxim A and Kovalev, Alexey K and Panov, Aleksandr I},
  journal={arXiv preprint arXiv:2508.09032},
  year={2025}
}

@article{bhat20253d,
  title={{3D CAVLA}: Leveraging Depth and {3D} Context to Generalize Vision Language Action Models for Unseen Tasks},
  author={Bhat, Vineet and Lan, Yu-Hsiang and Krishnamurthy, Prashanth and Karri, Ramesh and Khorrami, Farshad},
  journal={arXiv preprint arXiv:2505.05800},
  year={2025}
}

@article{li2025pointvla,
  title={{PointVLA}: Injecting the {3D} world into vision-language-action models},
  author={Li, Chengmeng and Wen, Junjie and Peng, Yan and Peng, Yaxin and Feng, Feifei and Zhu, Yichen},
  journal={arXiv preprint arXiv:2503.07511},
  year={2025}
}

@article{qu2025spatialvla,
  title={{SpatialVLA}: Exploring spatial representations for visual-language-action model},
  author={Qu, Delin and Song, Haoming and Chen, Qizhi and Yao, Yuanqi and Ye, Xinyi and Ding, Yan and Wang, Zhigang and Gu, JiaYuan and Zhao, Bin and Wang, Dong and others},
  journal={arXiv preprint arXiv:2501.15830},
  year={2025}
}

@article{zhen20243d,
  title={{3D-VLA}: A {3D} vision-language-action generative world model},
  author={Zhen, Haoyu and Qiu, Xiaowen and Chen, Peihao and Yang, Jincheng and Yan, Xin and Du, Yilun and Hong, Yining and Gan, Chuang},
  journal={arXiv preprint arXiv:2403.09631},
  year={2024}
}

@article{zhao2025vlas,
  title={{VLAS}: Vision-language-action model with speech instructions for customized robot manipulation},
  author={Zhao, Wei and Ding, Pengxiang and Zhang, Min and Gong, Zhefei and Bai, Shuanghao and Zhao, Han and Wang, Donglin},
  journal={arXiv preprint arXiv:2502.13508},
  year={2025}
}

@article{bi2025vla,
  title={{VLA-Touch}: Enhancing vision-language-action models with dual-level tactile feedback},
  author={Bi, Jianxin and Ma, Kevin Yuchen and Hao, Ce and Shou, Mike Zheng and Soh, Harold},
  journal={arXiv preprint arXiv:2507.17294},
  year={2025}
}

@article{huang2025tactile,
  title={{Tactile-VLA}: Unlocking Vision-Language-Action Model's Physical Knowledge for Tactile Generalization},
  author={Huang, Jialei and Wang, Shuo and Lin, Fanqi and Hu, Yihang and Wen, Chuan and Gao, Yang},
  journal={arXiv preprint arXiv:2507.09160},
  year={2025}
}

@article{wolters2024unleashing,
  title={Unleashing hydra: Hybrid fusion, depth consistency and radar for unified 3d perception},
  author={Wolters, Philipp and Gilg, Johannes and Teepe, Torben and Herzog, Fabian and Laouichi, Anouar and Hofmann, Martin and Rigoll, Gerhard},
  journal={arXiv preprint arXiv:2403.07746},
  year={2024}
}

@article{liu2022bevfusion,
  title={Bevfusion: Multi-task multi-sensor fusion with unified bird's-eye view representation},
  author={Liu, Zhijian and Tang, Haotian and Amini, Alexander and Yang, Xinyu and Mao, Huizi and Rus, Daniela and Han, Song},
  journal={arXiv preprint arXiv:2205.13542},
  year={2022}
}

@InProceedings{Wang_2024_CVPR,
    author    = {Wang, Tai and Mao, Xiaohan and Zhu, Chenming and Xu, Runsen and Lyu, Ruiyuan and Li, Peisen and Chen, Xiao and Zhang, Wenwei and Chen, Kai and Xue, Tianfan and Liu, Xihui and Lu, Cewu and Lin, Dahua and Pang, Jiangmiao},
    title     = {EmbodiedScan: A Holistic Multi-Modal 3D Perception Suite Towards Embodied AI},
    booktitle = {Proceedings of the IEEE/CVF Conference on Computer Vision and Pattern Recognition (CVPR)},
    month     = {June},
    year      = {2024},
    pages     = {19757-19767}
}

@InProceedings{Lin_2024_CVPR,
    author    = {Lin, Zhiwei and Liu, Zhe and Xia, Zhongyu and Wang, Xinhao and Wang, Yongtao and Qi, Shengxiang and Dong, Yang and Dong, Nan and Zhang, Le and Zhu, Ce},
    title     = {RCBEVDet: Radar-camera Fusion in Bird's Eye View for 3D Object Detection},
    booktitle = {Proceedings of the IEEE/CVF Conference on Computer Vision and Pattern Recognition (CVPR)},
    month     = {June},
    year      = {2024},
    pages     = {14928-14937}
}

@article{han2025multimodal,
  title={Multimodal fusion and vision-language models: A survey for robot vision},
  author={Han, Xiaofeng and Chen, Shunpeng and Fu, Zenghuang and Feng, Zhe and Fan, Lue and An, Dong and Wang, Changwei and Guo, Li and Meng, Weiliang and Zhang, Xiaopeng and others},
  journal={arXiv preprint arXiv:2504.02477},
  year={2025}
}

@article{xiong2025lxlv2,
  title={Lxlv2: Enhanced lidar excluded lean 3d object detection with fusion of 4d radar and camera},
  author={Xiong, Weiyi and Zou, Zean and Zhao, Qiuchi and He, Fengchun and Zhu, Bing},
  journal={IEEE Robotics and Automation Letters},
  year={2025},
  publisher={IEEE}
}

@article{kim2024crt,
  title={Crt-fusion: Camera, radar, temporal fusion using motion information for 3d object detection},
  author={Kim, Jisong and Seong, Minjae and Choi, Jun Won},
  journal={Advances in Neural Information Processing Systems},
  volume={37},
  pages={108625--108648},
  year={2024}
}

@inproceedings{li2025rctrans,
  title={Rctrans: Radar-camera transformer via radar densifier and sequential decoder for 3d object detection},
  author={Li, Yiheng and Yang, Yang and Lei, Zhen},
  booktitle={Proceedings of the AAAI Conference on Artificial Intelligence},
  volume={39},
  number={5},
  pages={5048--5056},
  year={2025}
}

@inproceedings{palladin2024samfusion,
  title={Samfusion: Sensor-adaptive multimodal fusion for 3d object detection in adverse weather},
  author={Palladin, Edoardo and Dietze, Roland and Narayanan, Praveen and Bijelic, Mario and Heide, Felix},
  booktitle={European Conference on Computer Vision},
  pages={484--503},
  year={2024},
  organization={Springer}
}

@article{zheng2025doracamom,
  title={Doracamom: Joint 3d detection and occupancy prediction with multi-view 4d radars and cameras for omnidirectional perception},
  author={Zheng, Lianqing and Liu, Jianan and Guan, Runwei and Yang, Long and Lu, Shouyi and Li, Yuanzhe and Bai, Xiaokai and Bai, Jie and Ma, Zhixiong and Shen, Hui-Liang and others},
  journal={arXiv preprint arXiv:2501.15394},
  year={2025}
}

@inproceedings{chebotar2023q,
  title={Q-transformer: Scalable offline reinforcement learning via autoregressive q-functions},
  author={Chebotar, Yevgen and Vuong, Quan and Hausman, Karol and Xia, Fei and Lu, Yao and Irpan, Alex and Kumar, Aviral and Yu, Tianhe and Herzog, Alexander and Pertsch, Karl and others},
  booktitle={Conference on Robot Learning},
  pages={3909--3928},
  year={2023},
  organization={PMLR}
}

@inproceedings{luo2023action,
  title={Action-quantized offline reinforcement learning for robotic skill learning},
  author={Luo, Jianlan and Dong, Perry and Wu, Jeffrey and Kumar, Aviral and Geng, Xinyang and Levine, Sergey},
  booktitle={Conference on Robot Learning},
  pages={1348--1361},
  year={2023},
  organization={PMLR}
}

@article{li2025vitra,
  title={Scalable vision-language-action model pretraining for robotic manipulation with real-life human activity videos},
  author={Li, Qixiu and Deng, Yu and Liang, Yaobo and Luo, Lin and Zhou, Lei and Yao, Chengtang and Zeng, Lingqi and Feng, Zhiyuan and Liang, Huizhi and Xu, Sicheng and others},
  journal={arXiv preprint arXiv:2510.21571},
  year={2025}
}

@article{steiner2024paligemma,
  title={Paligemma 2: A family of versatile vlms for transfer},
  author={Steiner, Andreas and Pinto, Andr{\'e} Susano and Tschannen, Michael and Keysers, Daniel and Wang, Xiao and Bitton, Yonatan and Gritsenko, Alexey and Minderer, Matthias and Sherbondy, Anthony and Long, Shangbang and others},
  journal={arXiv preprint arXiv:2412.03555},
  year={2024}
}

@inproceedings{zhai2023siglip,
  title={Sigmoid loss for language image pre-training},
  author={Zhai, Xiaohua and Mustafa, Basil and Kolesnikov, Alexander and Beyer, Lucas},
  booktitle={Proceedings of the IEEE/CVF international conference on computer vision},
  pages={11975--11986},
  year={2023}
}

@article{team2024gemma,
  title={Gemma 2: Improving open language models at a practical size},
  author={Team, Gemma and Riviere, Morgane and Pathak, Shreya and Sessa, Pier Giuseppe and Hardin, Cassidy and Bhupatiraju, Surya and Hussenot, L{\'e}onard and Mesnard, Thomas and Shahriari, Bobak and Ram{\'e}, Alexandre and others},
  journal={arXiv preprint arXiv:2408.00118},
  year={2024}
}

@inproceedings{peebles2023DiT,
  title={Scalable diffusion models with transformers},
  author={Peebles, William and Xie, Saining},
  booktitle={Proceedings of the IEEE/CVF international conference on computer vision},
  pages={4195--4205},
  year={2023}
}

@article{carion2025sam3,
  title={{SAM 3}: Segment anything with concepts},
  author={Carion, Nicolas and Gustafson, Laura and Hu, Yuan-Ting and Debnath, Shoubhik and Hu, Ronghang and Suris, Didac and Ryali, Chaitanya and Alwala, Kalyan Vasudev and Khedr, Haitham and Huang, Andrew and others},
  journal={arXiv preprint arXiv:2511.16719},
  year={2025}
}

@article{liu2025mla,
  title={{MLA}: A multisensory language-action model for multimodal understanding and forecasting in robotic manipulation},
  author={Liu, Zhuoyang and Liu, Jiaming and Xu, Jiadong and Han, Nuowei and Gu, Chenyang and Chen, Hao and Zhou, Kaichen and Zhang, Renrui and Hsieh, Kai Chin and Wu, Kun and others},
  journal={arXiv preprint arXiv:2509.26642},
  year={2025}
}

@InProceedings{pmlr-pi05,
  title={$\pi_{0.5}$: a Vision-Language-Action Model with Open-World Generalization},
  author={Black, Kevin and Brown, Noah and Darpinian, James and Dhabalia, Karan and Driess, Danny and Esmail, Adnan and Equi, Michael Robert and Finn, Chelsea and Fusai, Niccolo and Galliker, Manuel Y. and Ghosh, Dibya and Groom, Lachy and Hausman, Karol and Ichter, Brian and Jakubczak, Szymon and Jones, Tim and Ke, Liyiming and LeBlanc, Devin and Levine, Sergey and Li-Bell, Adrian and Mothukuri, Mohith and Nair, Suraj and Pertsch, Karl and Ren, Allen Z. and Shi, Lucy Xiaoyang and Smith, Laura and Springenberg, Jost Tobias and Stachowicz, Kyle and Tanner, James and Vuong, Quan and Walke, Homer and Walling, Anna and Wang, Haohuan and Yu, Lili and Zhilinsky, Ury},
  booktitle = 	 {Proceedings of The 9th Conference on Robot Learning},
  pages = 	 {17--40},
  year = 	 {2025},
  editor = 	 {Lim, Joseph and Song, Shuran and Park, Hae-Won},
  volume = 	 {305},
  series = 	 {Proceedings of Machine Learning Research},
  month = 	 {27--30 Sep},
  publisher =    {PMLR},
  url = 	 {https://proceedings.mlr.press/v305/black25a.html},
}

@misc{molmoact2025,
      title={{MolmoAct}: Action Reasoning Models that can Reason in Space},
      author={Jason Lee and Jiafei Duan and Haoquan Fang and Yuquan Deng and Shuo Liu and Boyang Li and Bohan Fang and Jieyu Zhang and Yi Ru Wang and Sangho Lee and Winson Han and Wilbert Pumacay and Angelica Wu and Rose Hendrix and Karen Farley and Eli VanderBilt and Ali Farhadi and Dieter Fox and Ranjay Krishna},
      year={2025},
      eprint={2508.07917},
      archivePrefix={arXiv},
      primaryClass={cs.RO},
      url={https://arxiv.org/abs/2508.07917}
}

@misc{contributors2024agibotworldrepo,
  title={AgiBot World Colosseum},
  author={AgiBot World Colosseum contributors},
  howpublished={\url{https://github.com/OpenDriveLab/AgiBot-World}},
  year={2024}
}

@inproceedings{jones2025FuSe,
  title={Beyond sight: Finetuning generalist robot policies with heterogeneous sensors via language grounding},
  author={Jones, Joshua and Mees, Oier and Sferrazza, Carmelo and Stachowicz, Kyle and Abbeel, Pieter and Levine, Sergey},
  booktitle={2025 IEEE International Conference on Robotics and Automation (ICRA)},
  pages={5961--5968},
  year={2025},
  organization={IEEE}
}

@article{yu2025forcevla,
  title={Forcevla: Enhancing vla models with a force-aware moe for contact-rich manipulation},
  author={Yu, Jiawen and Liu, Hairuo and Yu, Qiaojun and Ren, Jieji and Hao, Ce and Ding, Haitong and Huang, Guangyu and Huang, Guofan and Song, Yan and Cai, Panpan and others},
  journal={arXiv preprint arXiv:2505.22159},
  year={2025}
}

@article{wu2026pragmatic,
  title={A Pragmatic VLA Foundation Model},
  author={Wei Wu and Fan Lu and Yunnan Wang and Shuai Yang and Shi Liu and Fangjing Wang and Shuailei Ma and He Sun and Yong Wang and Zhenqi Qiu and Houlong Xiong and Ziyu Wang and Shuai Zhou and Yiyu Ren and Kejia Zhang and Hui Yu and Jingmei Zhao and Qian Zhu and Ran Cheng and Yong-Lu Li and Yongtao Huang and Xing Zhu and Yujun Shen and Kecheng Zheng},
  journal={arXiv preprint arXiv:2601.18692v1},
  year={2026}
}

@article{intelligence2026pi,
  title={$\pi_{0.7}$: A Steerable Generalist Robotic Foundation Model with Emergent Capabilities},
  author={Intelligence, Physical and Ai, Bo and Amin, Ali and Aniceto, Raichelle and Balakrishna, Ashwin and Balke, Greg and Black, Kevin and Bokinsky, George and Cao, Shihao and Charbonnier, Thomas and others},
  journal={arXiv preprint arXiv:2604.15483},
  year={2026}
}

@article{bjorck2025gr00t,
  title={Gr00t n1: An open foundation model for generalist humanoid robots},
  author={Bjorck, Johan and Casta{\~n}eda, Fernando and Cherniadev, Nikita and Da, Xingye and Ding, Runyu and Fan, Linxi and Fang, Yu and Fox, Dieter and Hu, Fengyuan and Huang, Spencer and others},
  journal={arXiv preprint arXiv:2503.14734},
  year={2025}
}

@article{generalist2026gen1,
author = {Generalist AI Team},
title = {GEN-1: Scaling Embodied Foundation Models to Mastery},
journal = {Generalist AI Blog},
year = {2026},
note = {https://generalistai.com/blog/apr-02-2026-GEN-1},
}

@article{guo2025omnivla,
  title={OmniVLA: Physically-Grounded Multimodal VLA with Unified Multi-Sensor Perception for Robotic Manipulation},
  author={Guo, Heyu and Wang, Shanmu and Ma, Ruichun and Jiang, Shiqi and Ghasempour, Yasaman and Abari, Omid and Guo, Baining and Qiu, Lili},
  journal={arXiv preprint arXiv:2511.01210},
  year={2025}
}

@article{gao2026dreamdojo,
  title={DreamDojo: A Generalist Robot World Model from Large-Scale Human Videos},
  author={Gao, Shenyuan and Liang, William and Zheng, Kaiyuan and Malik, Ayaan and Ye, Seonghyeon and Yu, Sihyun and Tseng, Wei-Cheng and Dong, Yuzhu and Mo, Kaichun and Lin, Chen-Hsuan and others},
  journal={arXiv preprint arXiv:2602.06949},
  year={2026}
}

@article{zhang2026clap,
  title={CLAP: Contrastive Latent Action Pretraining for Learning Vision-Language-Action Models from Human Videos},
  author={Zhang, Chubin and Wang, Jianan and Gao, Zifeng and Su, Yue and Dai, Tianru and Zhou, Cai and Lu, Jiwen and Tang, Yansong},
  journal={arXiv preprint arXiv:2601.04061},
  year={2026}
}
